%% file: main.tex
\PassOptionsToPackage{most,breakable}{tcolorbox}
\PassOptionsToPackage{dvipsnames,table}{xcolor}
\PassOptionsToPackage{numbers,compress}{natbib}
\documentclass[11pt]{article}

\usepackage[final]{ropedia}

\usepackage[T1]{fontenc}
\usepackage[utf8]{inputenc}
\usepackage{amsmath,amssymb}
\usepackage{amsfonts}
\usepackage{mathpazo}
\usepackage[xcharter,bigdelims,vvarbb]{newtxmath}
\usepackage[scaled]{helvet}
\usepackage[scaled=1.1]{zlmtt}
\usepackage{latexsym}
\usepackage{makecell}
\usepackage{tabularx}
\usepackage{pifont}
\usepackage{placeins}
\usepackage{booktabs}
\usepackage{multirow}
\usepackage{tabularx}
\usepackage{subcaption}
\usepackage{adjustbox}
\usepackage{makecell}
\usepackage{enumitem}
\usepackage{rotating}
\usepackage{titletoc}
\usepackage{comment}
\usepackage{wrapfig}
\usepackage{needspace}
\usepackage{nicefrac}
\usepackage{microtype}
\usepackage{siunitx}
\usepackage{dblfloatfix}
\usepackage{graphicx}
\usepackage{cuted}
\usepackage{capt-of}
\usepackage{xspace}
\setlist[itemize]{leftmargin=2em}


\input{macro}

\firstpagelogos{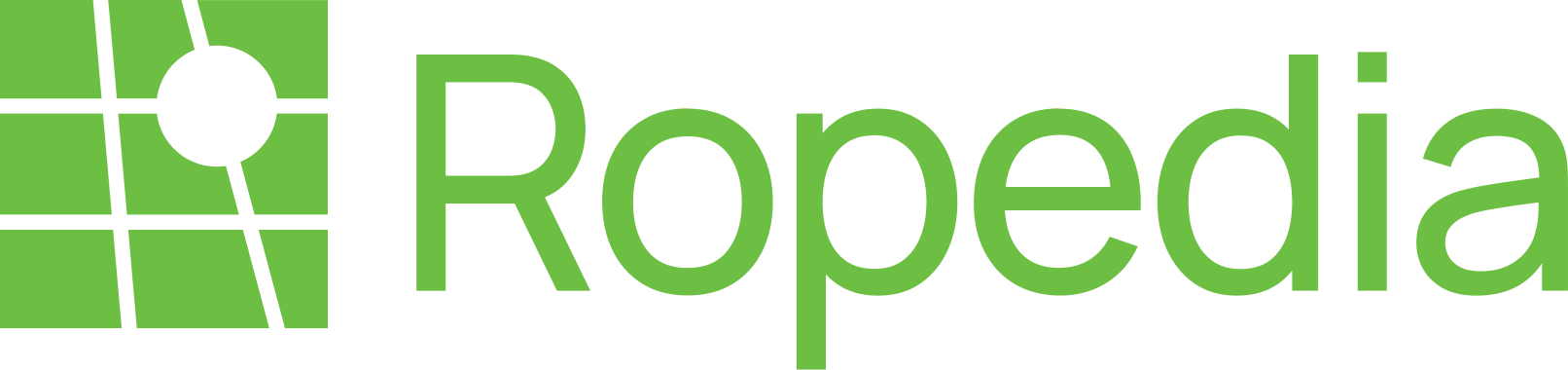}{}{}
\abstractlogo{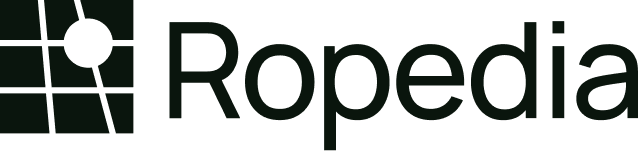}

\title{\projecttitle}
\subtitle{\projectsubtitle}

\author{\textbf{Ropedia Research Team}}

\newcommand{\teaserplaceholder}{%
  \begin{tcolorbox}[
      colback=ropediaabstractgreen,
      colframe=ropediaaccent,
      coltext=ropediaink,
      boxrule=0.6pt,
      width=\textwidth,
      height=0.32\textwidth,
      valign=center,
      arc=2mm,
      left=6mm,
      right=6mm
    ]
    \centering\sffamily
    {\Large\bfseries Teaser Placeholder}\\[0.6ex]
    Replace this box by adding \texttt{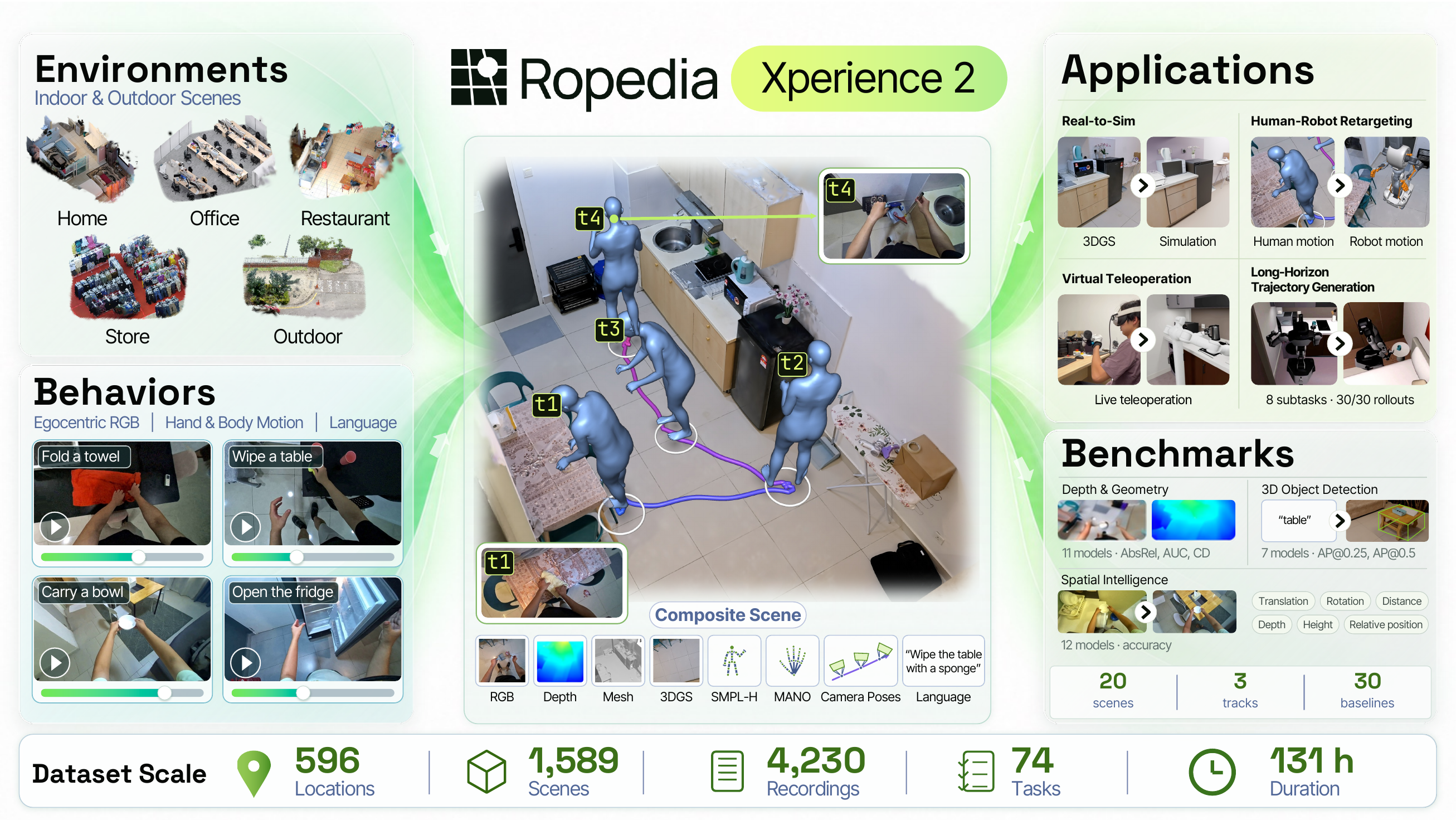}.
  \end{tcolorbox}%
}

\begin{document}
\maketitle

\begin{paperresources}
\paperresourceicon{\resourceprojecticon}{Website}{https://ropedia.com/xperience/2-research/}{ropedia.com/xperience/2-research/}\\[1pt]
\paperresourceicon{\resourcegithubicon}{GitHub}{https://github.com/Ropedia/Ropedia-Xperience-2.git}{github.com/Ropedia/Ropedia-Xperience-2.git}\\[1pt]
\paperresourceicon{\resourcehficon}{Hugging Face}{https://huggingface.co/datasets/ropedia-ai/Ropedia-Xperience-2}{huggingface.co/datasets/ropedia-ai/Ropedia-Xperience-2}
\end{paperresources}

\vspace{6pt}
{\centering
\IfFileExists{figs/teaser.pdf}{%
  \includegraphics[width=1\textwidth]{figs/teaser.pdf}%
}{%
  \teaserplaceholder
}
\par\vspace{5pt}
\begin{minipage}{\textwidth}
\footnotesize
\refstepcounter{figure}\label{fig:teaser}
Figure~\thefigure{} | \textbf{\datasetname overview.}
The pipeline can register egocentric HOMIE recordings to 3D-scanned environments (left), bringing body motion, camera poses, and language into a shared scene representation (center) to support downstream applications and benchmarks (right).
The virtual teleoperation panel shows the operator and a simulated robot manipulating a kettle.
The scene count refers to released scans; location, recording, task, and duration counts describe the broader collection inventory.
\end{minipage}
\par}
\vspace{6pt}

\begin{abstract}
\input{secs/00_abstract}
\end{abstract}

\input{body}

\end{document}

%% file: macro.tex
\usepackage{tikz}
\usetikzlibrary{fadings}
\usepackage{array}
\usepackage{tcolorbox}
\usepackage{threeparttable}

\newcommand{\projecttitle}{Ropedia Xperience 2\xspace}
\newcommand{\projectsubtitle}{A Gold Standard for Scaling Physical AI Environments and Behaviors\xspace}
\newcommand{\datasetname}{Ropedia Xperience 2\xspace}

\newcommand{\homie}{HOMIE\xspace}

\input{tabs/release_numbers}

\newcommand{\cmark}{\textcolor{green!60!black}{\ding{51}}}

\definecolor{subcol}{HTML}{F7FFF2}
\colorlet{rankfirst}{ropediagreen}
\colorlet{ranksecond}{ropedialightgreen}
\colorlet{rankthird}{ropediaabstractgreen}
\colorlet{bestgreen}{rankfirst}
\colorlet{secondgreen}{ranksecond}
\colorlet{thirdgreen}{rankthird}
\colorlet{bestred}{rankfirst}
\colorlet{secondorange}{ranksecond}
\colorlet{thirdyellow}{rankthird}
\definecolor{trainmark}{RGB}{0,180,0}
\definecolor{oomred}{RGB}{252,228,228}
\definecolor{catgray}{HTML}{E7F8DC}
\definecolor{altcolgreen}{HTML}{F0FFE8}
\colorlet{altcolblue}{altcolgreen}
\definecolor{pretrainmark}{RGB}{197,90,17}
\definecolor{gaingreen}{RGB}{30,140,60}
\definecolor{lossred}{RGB}{200,30,30}
\definecolor{navyblue}{HTML}{4F8F35}

\colorlet{findingback}{ropedialightgreen}
\colorlet{findingbacksoft}{ropediaabstractgreen}
\colorlet{findingframe}{ropediadeepgreen}
\colorlet{findingaccent}{ropediaaccent}


%% file: tabs/release_numbers.tex
\newcommand{\ReleaseScenes}{1{,}589}
\newcommand{\ReleasePerfect}{148}
\newcommand{\ReleaseUsable}{934}
\newcommand{\ReleaseReuse}{507}
\newcommand{\ReleaseDelete}{21}
\newcommand{\ReleaseDecided}{1{,}610}
\newcommand{\ReleaseUngraded}{168}
\newcommand{\ReleaseIncomplete}{2}
\newcommand{\ReleaseUploadWorks}{1{,}780}
\newcommand{\ReleaseHomeScenes}{1{,}389}
\newcommand{\ReleaseOutdoorScenes}{152}

\newcommand{\ReleaseCorridorPerfect}{42}

\newcommand{\CollectionLocations}{596}
\newcommand{\CollectionHomes}{443}
\newcommand{\CollectionTasks}{74}
\newcommand{\CollectionRecordings}{4{,}230}
\newcommand{\CollectionHours}{131}
\newcommand{\ReleaseDeleteDoors}{12}
\newcommand{\ReleaseDeleteFailed}{8}
\newcommand{\ReleaseDeletePeople}{1}
\newcommand{\ReleaseSupplementUsable}{84}
\newcommand{\ReleaseOverrideUsable}{7}
\newcommand{\ReleaseRuleUsable}{91}
\newcommand{\ReleaseConflicts}{33}
\newcommand{\ReleasePerfectPercent}{9.3\%}
\newcommand{\ReleaseUsablePercent}{58.8\%}
\newcommand{\ReleaseReusePercent}{31.9\%}
\newcommand{\ReleaseRetainedPercent}{98.7\%}
\newcommand{\ReleaseDeletePercent}{1.3\%}

\newcommand{\ReleaseHomePercent}{87.4\%}
\newcommand{\ReleaseBedroomLivingPercent}{61.4\%}
\newcommand{\ReleaseKitchenReusePercent}{81.6\%}
\newcommand{\ReleaseBathroomReusePercent}{58.0\%}
\newcommand{\ReleaseOutdoorGoodPercent}{86.2\%}

%% file: secs/00_abstract.tex
Large foundation models such as GPT-6 Astra and Claude Fable 5.1 have recently been shown to control robots directly from visual observations.
They nevertheless remain brittle on long-horizon, contact-rich tasks in real-world environments.
Further progress depends on two complementary resources: demonstrations of human behavior that supervise how tasks are performed, and interactive environments in which policies can be rolled out, evaluated, and improved in closed loop.
Language models have advanced along well-characterized scaling laws; the corresponding question for physical AI is how capability scales with environments and behaviors, which first requires identifying which data to scale and how to produce and reuse such data continuously.
To address this question, we introduce \textbf{\datasetname}, comprising \ReleaseScenes{} released metric-scale 3D scans and a collection of \CollectionRecordings{} egocentric recordings of \CollectionTasks{} tasks (\CollectionHours{} hours).
Data are captured with a handheld LiDAR scanner and a head-mounted multi-camera recorder, without robots or motion-capture infrastructure; a processing pipeline reconstructs meshes and 3D Gaussian splatting models, estimates camera trajectories and hand and body motion, generates hierarchical language annotations, and supports registration of recordings to scene geometry.
The reconstructions serve as reference for benchmarks on depth and geometry estimation, 3D object detection, and spatial reasoning.
We further show that each capture can be repurposed for robot learning: scenes are converted into physically interactive simulation environments, human motion is retargeted to whole-body robot trajectories, and additional robot behaviors are obtained through virtual teleoperation and automated long-horizon trajectory generation.
We propose this joint capture of environments and behaviors as a gold standard for physical AI data collection, laying a data foundation for establishing scaling laws over environments and behaviors.

%% file: body.tex
\input{secs/01_introduction}
\input{secs/03_data_collection}
\input{secs/04_processing_annotation}
\input{secs/05_dataset_statistics_quality}
\input{secs/06_applications}
\input{secs/07_release_limitations}
\input{secs/08_conclusion}

{
\small
\bibliography{references}
}

\par\bigskip
\appendix
\fontsize{10.5}{13}\selectfont
\setlength{\parskip}{2pt plus 1pt minus 0.5pt}
\captionsetup[table]{skip=5pt}
  \startcontents[appendix]
  \begingroup
  \small
  \setlength{\parskip}{0pt}
  \printcontents[appendix]{l}{1}{%
    \section*{Appendix Contents}%
    \vspace{-0.5em}%
  }
  \endgroup
  \medskip

\input{secs/appendix/a_contributors}
\input{secs/appendix/b_processing_annotation}
\input{secs/appendix/c_additional_statistics}
\input{secs/appendix/d_experimental_details}
\input{secs/appendix/e_acknowledgments}


%% file: secs/01_introduction.tex
\section{Introduction}
\label{sec:introduction}



Large foundation models such as GPT-6 Astra and Claude Fable 5.1 have recently been shown to control robots directly from visual observations~\citep{zhang2026unexpected,sharma2026roboarcade}.
On the 42-task RoboDojo benchmark~\citep{chen2026robodojo}, GPT-6 Astra attains a reported average success rate of 22.48\%, surpassing the 40 public policies in that comparison~\citep{zhang2026unexpected}, and it reaches 92\% on RxR~\citep{ku2020rxr} instruction following and 82\% on HM3D~\citep{ramakrishnan2021hm3d} object search~\citep{galbot2026astra}.
These models nevertheless remain far from practical deployment: an evaluation across eight robot configurations reports no success in 11 plug-insertion trials across three platforms and 9 successes in 23 navigation-and-search trials for GPT-6 Astra, while tracking progress across many subtasks remains difficult~\citep{sensenova2026astra}.

Enabling large foundation models to control robots reliably over long-horizon, complex tasks in real environments requires two complementary resources: (i) behavior demonstrations from which the models can learn, and (ii) environments in which they can execute tasks and accumulate experience.
Language models have advanced along well-characterized scaling laws relating performance to data, model size, and compute~\citep{kaplan2020scaling}; the corresponding question for physical AI is how capability scales with environments and behaviors.
Answering it first requires identifying which environment and behavior data to scale, and how such data can be produced and reused continuously.

\begin{figure}[t]
\centering
\includegraphics[width=\linewidth]{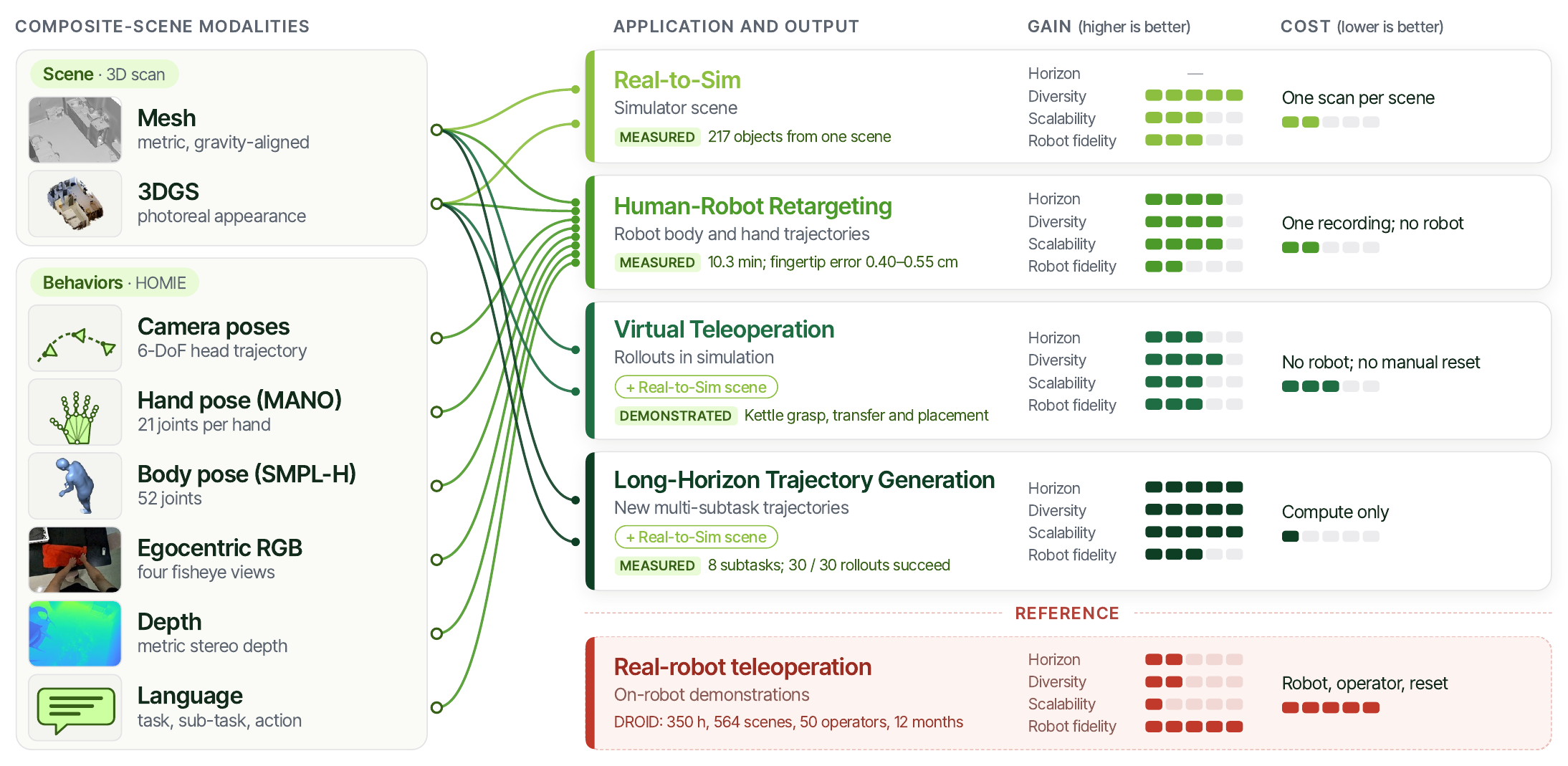}
\caption{\textbf{From composite-scene modalities to robot data.}
\emph{Left:} scene assets from the 3D scan and \homie behavior streams; lines link each modality to the applications that use it.
\emph{Right:} the output, gains, and cost per collection of each application, with real-robot teleoperation as the reference; chips mark applications that build on Real-to-Sim scenes.
Bars are qualitative five-step ratings, not measurements; the measured and demonstrated notes refer to results from Section~\ref{sec:application_demos}, and the reference row quotes DROID~\citep{khazatsky2024droid}.}
\label{fig:application_value}
\end{figure}

To address this question, we introduce \textbf{\datasetname}, a dataset of real environments and behaviors for physical AI.
\datasetname combines LiDAR scans of real scenes with egocentric recordings of human activity; our registration pipeline can place the corresponding scene geometry, video, camera trajectories, and hand and body motion in a common metric frame.
Registered recordings can support human-to-robot motion retargeting, while scanned scenes can be converted into interactive simulation environments; registration checks and per-scene quality control provide tools for auditing, filtering, and reusing the data.
With \datasetname, we aim to establish a gold standard for next-generation physical AI data collection and to lay the data foundation for scaling laws of physical AI over environments and behaviors.

\datasetname is collected with a two-stage protocol that requires a single operator and no robot or motion-capture infrastructure on site.
Each scene is first scanned once with a handheld 3D scanner, and its assigned tasks are then recorded with the \homie head-mounted recorder, so that one scan provides the spatial context for multiple recordings.
The released scan component contains \ReleaseScenes{} scenes.
The broader collection inventory spans \CollectionLocations{} locations and \CollectionRecordings{} task recordings totaling \CollectionHours{} hours.
An automated pipeline produces scene meshes and 3D Gaussian representations, alongside camera trajectories, depth, hand and body motion, and hierarchical language captions.
\datasetname extends Xperience-10M~\citep{xperience_10m}, which provides about 10 million egocentric interaction clips totaling 10,000 hours, with metric scene scans and a pipeline for registering egocentric recordings to them.

To assess the fidelity of the data, we use the scan reconstructions as reference geometry in three benchmarks: depth and geometry estimation, 3D object detection, and spatial reasoning.
These benchmarks evaluate existing models across diverse real-world scenes.

Each capture is further reused to scale both environments and behaviors for robot learning (Figure~\ref{fig:application_value}).
Here, \emph{environments} are real spaces with their layouts and object relations, and \emph{behaviors} comprise the recorded human task executions as well as the robot motions and interactions derived from them.
We demonstrate four routes:
\begin{itemize}
  \item \textbf{Real-to-sim conversion} turns a single scan of a real-world environment into an interactive digital twin that preserves the original layout and appearance, incorporates realistic physical properties, and supports part-level object interaction, yielding reusable environments for policy and control learning.
  \item \textbf{Human-robot retargeting} converts the recorded human behaviors, including long-horizon activities, into robot demonstrations from the estimated body and hand motion, without on-site robot deployment.
  \item \textbf{Virtual teleoperation} lets operators collect robot demonstrations in the converted simulation scenes, without physical robots and with automatic scene resets.
  \item \textbf{Long-horizon trajectory generation} rolls out robots on new multi-step tasks in the converted simulation scenes, chaining navigation and manipulation across subtasks from varied start states and object configurations; more trajectories require only additional compute.
\end{itemize}
\noindent
Real-to-sim conversion scales environments, while retargeting, virtual teleoperation, and trajectory generation scale robot behaviors.
Capture thus scales with operators rather than robots, and the captured data can be expanded into robot data without further on-site collection.
These routes complement real-robot teleoperation by offering lower-cost paths to training environments and demonstrations at different levels of fidelity.

%% file: secs/03_data_collection.tex
\section{Data Collection}
\label{sec:data_collection}


\begin{figure}[t]
  \centering
  \includegraphics[width=\linewidth]{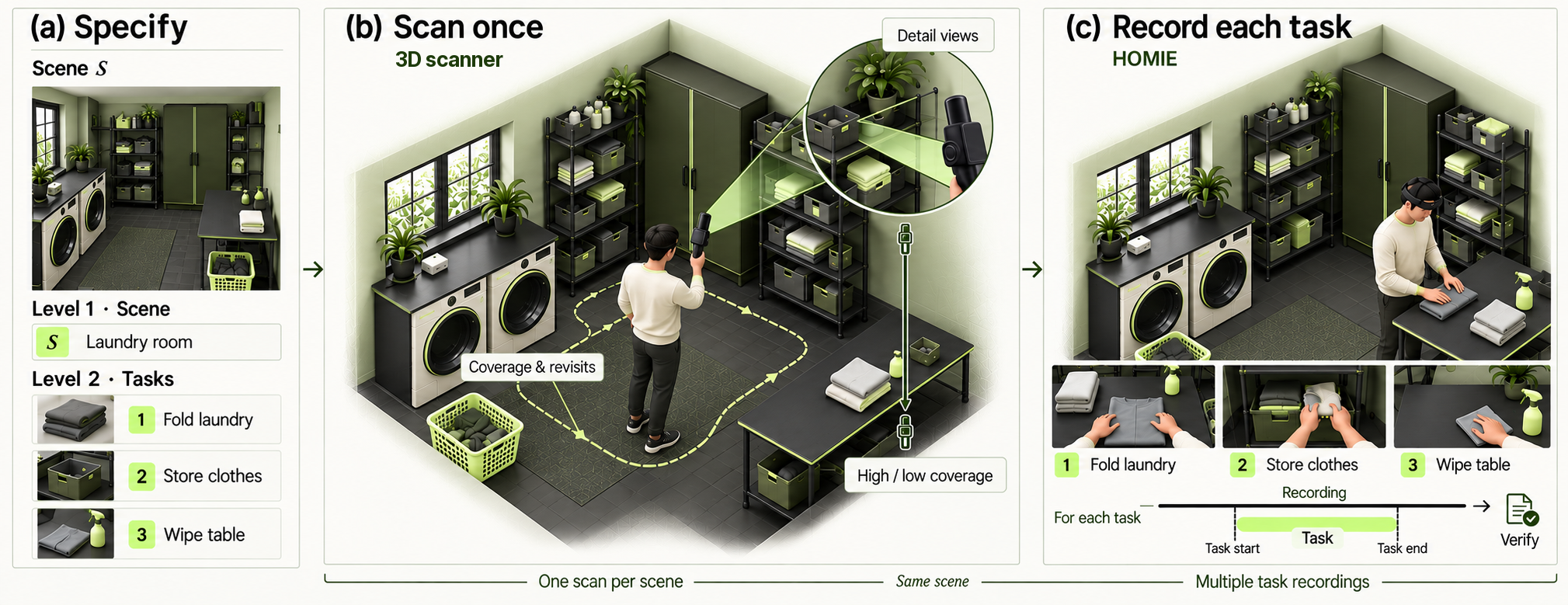}
  \caption{\textbf{Overview of data collection.}
  (a) Before capture, each scene (Level~1) is assigned a set of tasks (Level~2).
  (b) The scene is scanned once with the 3D scanner, with overlapping coverage, revisits, detail views, and high and low viewpoints.
  (c) Its tasks are then recorded individually with \homie; each recording begins before the task and ends after it, and the saved file is verified.
  One scene scan provides the environmental context for multiple task recordings.
  Rooms, tasks, viewpoints, and device views are illustrative.}
  \label{fig:collection_workflow}
\end{figure}

Our goal is to capture natural human activity together with the 3D structure of the environment in which it occurs, providing spatial context for understanding how people interact with their surroundings.
Scene collections such as ScanNet++ and DL3DV-10K emphasize multi-view observations of real environments~\citep{yeshwanth2023scannetpp,ling2024dl3dv}, while Ego4D and Nymeria capture egocentric activity and motion~\citep{grauman2022ego4d,ma2024nymeria}.
Holi-World organizes human-egocentric experience into grounded learning resources~\citep{li2026holiworld}, and SpatialBench evaluates spatial foundation models across scene domains and input densities~\citep{peng2026spatialbench}.
UMI further demonstrates portable human-demonstration capture without requiring a robot at the collection site~\citep{chi2024umi}.

We collect \datasetname in real, in-use environments, including homes, restaurants, stores, offices, and outdoor sites.
Collection is organized by scene: each scene (Level~1) is assigned a set of tasks (Level~2) before capture (Figure~\ref{fig:collection_workflow}a).

Capturing environmental geometry and natural activity requires different sensing configurations.
Scene scanning benefits from freely chosen viewpoints to observe high, low, and partially occluded regions, while activity recording requires both hands to remain free and the cameras to follow the operator's viewpoint.
We therefore use two complementary devices (Table~\ref{tab:capture_devices}).
The 3D scanner is a handheld LiDAR device equipped with two fisheye and two wide-angle cameras, allowing the operator to scan the scene from varied viewpoints.
\homie is a head-mounted egocentric recorder with four global-shutter fisheye cameras, two facing forward and two facing backward.
The forward pair captures the hands and manipulated objects, while the global shutters avoid rolling-shutter distortion during rapid head motion.
\homie also records inertial measurements and four-channel audio.

For each scene, the operator first performs a 3D scan and then wears \homie to complete the assigned tasks one by one (Figure~\ref{fig:collection_workflow}).
The protocol is designed to associate task recordings with a 3D scan of the same scene, linking activity to its physical environment; Section~\ref{sec:spatial_registration} describes how they can be spatially registered.
The following subsections describe the scenes and tasks and the capture protocol.

\input{tabs/capture_devices}
\FloatBarrier

\subsection{Scenes and Tasks}
\label{sec:scenes_tasks}

\begin{figure}[!htbp]
  \centering
  \includegraphics[width=\linewidth]{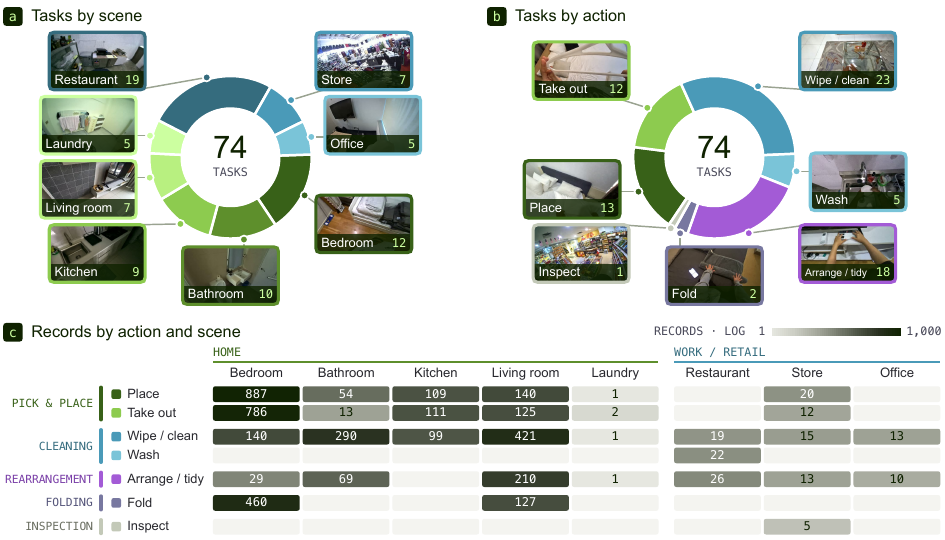}
  \caption{\textbf{Tasks by scene and action in the collection inventory.}
  (a, b) Number of tasks per scene type and per action; each thumbnail is a representative \homie frame of that category.
  (c) Number of uploaded recordings per action and scene type, with log-scaled color; counts do not require completed processing or registration.}
  \label{fig:task_statistics}
\end{figure}

The collection inventory contains \CollectionTasks{} Level~2 tasks, 31 of which take place in work and retail scenes such as restaurants, stores, and offices (Figure~\ref{fig:task_statistics}a).
By leading verb, the tasks fall into seven actions, which form five categories: pick and place, cleaning, rearrangement, folding, and inspection (Figure~\ref{fig:task_statistics}b).
Placing or taking out objects (25 tasks), wiping and cleaning (23), and arranging and tidying (18) account for most tasks.
The \CollectionRecordings{} uploaded recordings, each one execution of a task, are concentrated in home scenes: placing, taking out, and folding dominate in bedrooms, and wiping in living rooms and bathrooms (Figure~\ref{fig:task_statistics}c).

\FloatBarrier

\subsection{Capture Protocol}
\label{sec:capture_protocol}

We standardize collection with a written protocol so that captures are consistent across sites and operators.
It covers scene preparation, 3D scanning, and \homie task recording; the latter two are shown in Figure~\ref{fig:collection_workflow}b and c.

\paragraph{Scene preparation.}
The scene must remain static from the start of the 3D scan until task recording begins.
Before capture, we therefore minimize appearance changes and transient content: screens are turned off or set to a static image, lights are turned on and kept constant, bystanders and pets leave the scene, and windows and other transparent surfaces are covered where possible.

\paragraph{3D scanning.}
Scanning aims for complete coverage with redundant views.
The operator walks slowly through the scene, traversing each route in both directions, scans occluded and detailed regions at close range, and sweeps the device vertically to add high and low viewpoints.

\paragraph{\homie task recording.}
Task recording aims to capture each activity completely and naturally.
Before each recording, the operator checks that the device is secure and that no lens is obstructed.
Recording begins before the task starts and ends after it is completed, so that each recording contains the entire activity.

\FloatBarrier

%% file: tabs/capture_devices.tex
\begin{table}[!htbp]
\centering
\caption{\textbf{Capture devices.}
The 3D scanner captures the static environment, and \homie records the operator's activity.
A dash indicates that the sensor is absent.}
\label{tab:capture_devices}
\begingroup
\small
\setlength{\tabcolsep}{3pt}
\renewcommand{\arraystretch}{1.18}
\begin{tabularx}{\textwidth}{@{}l*{5}{>{\centering\arraybackslash}X}c@{}}
\toprule
\textbf{Device} & \textbf{Form factor} & \textbf{Cameras} & \textbf{Shutter} & \textbf{LiDAR} & \textbf{IMU} & \textbf{Audio channels} \\
\midrule
\textbf{3D scanner} & Handheld & \makecell[c]{2 fisheye,\\2 wide-angle} & Rolling & \cmark & \textemdash & \textemdash \\
\addlinespace[3pt]
\textbf{\homie} & Head-mounted & \makecell[c]{4 fisheye\\(2 front, 2 rear)} & Global & \textemdash & \cmark & 4 \\
\bottomrule
\end{tabularx}
\endgroup
\end{table}

%% file: secs/04_processing_annotation.tex
\section{Processing and Annotation}
\label{sec:processing_annotation}

\begin{figure}[tbp]
  \centering
  \includegraphics[width=\linewidth]{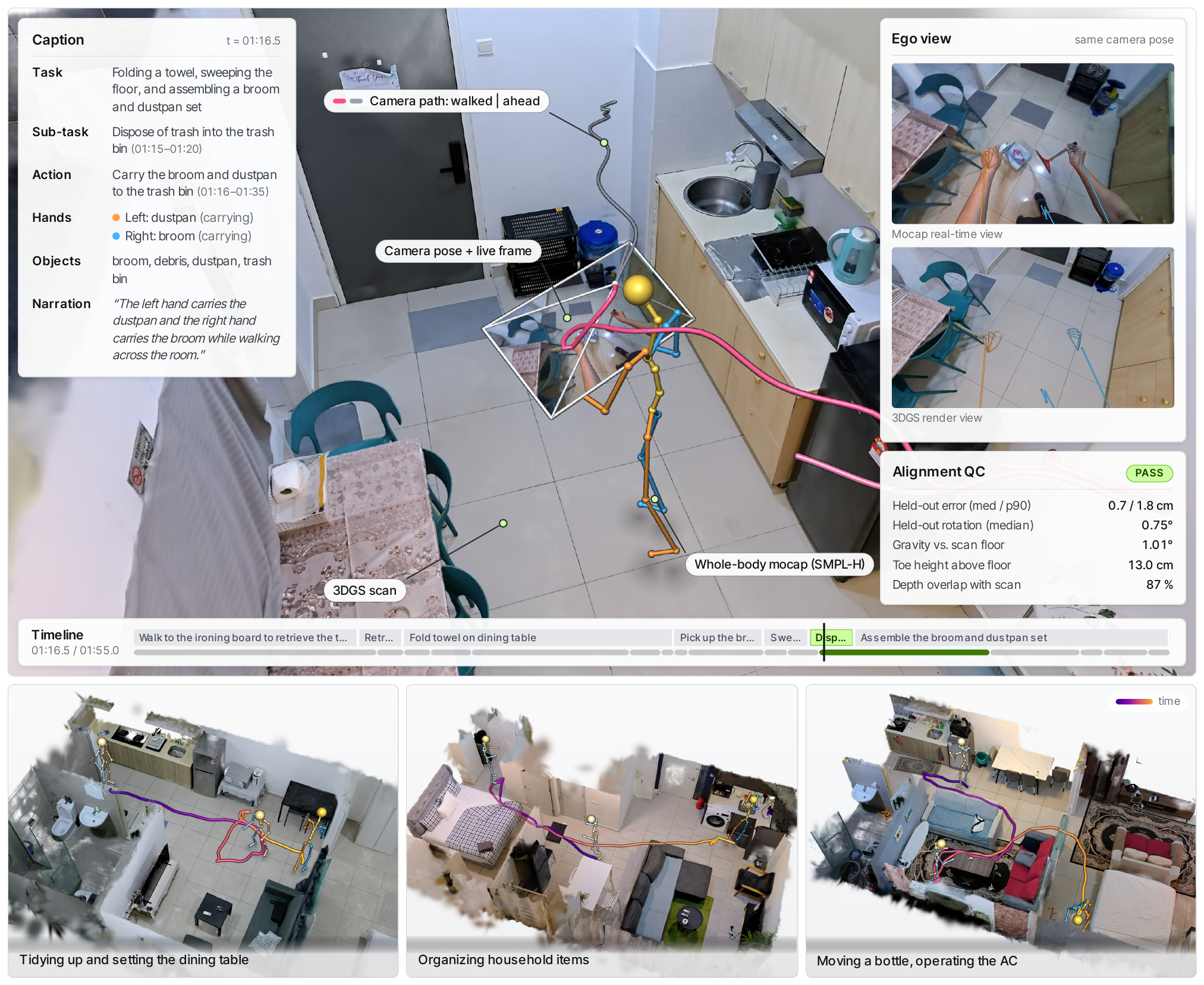}
  \caption{\textbf{Egocentric recordings registered into 3D Gaussian scene scans.}
  \emph{Top:} synchronized scene, motion, video, and activity annotation views of one recording.
  Orange/blue denote the body's left/right sides; pink/grey show the past 20\,s/next 12\,s of the camera path.
  \emph{Bottom:} 45\,s motion excerpts from three further scenes, with ground tracks coloured by time and poses at the start, middle, and end of each window (earlier poses in pale colours).}
  \label{fig:viewer}
\end{figure}

For each scene, the 3D scan provides a static, metric 3D reconstruction (Section~\ref{sec:geometric_processing}).
The \homie processing pipeline can turn task recordings into synchronized video, a device trajectory, and hand, body, and activity annotations (Section~\ref{sec:video_processing}).
The pipeline can register a recording's device trajectory and estimated hand and body motion to the reconstruction of the corresponding scene (Section~\ref{sec:spatial_registration}).
Table~\ref{tab:asset_provenance} lists the resulting assets and their provenance, and Figure~\ref{fig:viewer} shows example recordings registered into their scenes.

\input{tabs/asset_provenance}

\subsection{3D Scan Processing}
\label{sec:geometric_processing}

The 3D scan exports comprise fisheye images with their camera poses and a sparse point model in COLMAP format~\citep{schoenberger2016sfm}, a triangle mesh, and a 3D Gaussian splatting (3DGS) model~\citep{kerbl20233dgs}; available assets depend on the export.
These assets share one coordinate frame, which we call the \emph{scene frame}; because the geometry is measured by LiDAR, the scene frame is metric.
The posed images and the mesh serve as the reference for registering the task recordings of the scene (Section~\ref{sec:spatial_registration}), and the 3DGS model renders the scene from arbitrary viewpoints, such as the recorded \homie camera poses (Figure~\ref{fig:viewer}).

\subsection{\texorpdfstring{\homie}{HOMIE} Processing}
\label{sec:video_processing}

We use the same automated pipeline to process \homie recordings.

\paragraph{Synchronization and calibration.}
The four cameras are hardware-synchronized, and all sensor samples are timestamped by the device clock.
We place the four videos on a common 30\,fps frame index and link the IMU and audio samples to it by timestamp.
Each device is calibrated before capture, and every recording stores the intrinsics of the four fisheye cameras and their extrinsics relative to the IMU (Appendix~\ref{appendix:sync_calibration}).

\paragraph{Device trajectory and depth.}
The two forward-facing views are rectified into a pinhole stereo pair, on which SLAM, depth, and hand estimation operate.
SLAM estimates a 6-DoF device pose for every frame and a sparse landmark map in a gravity-aligned metric frame whose origin is the head position at the first frame.
A stereo module additionally estimates metric depth with per-pixel confidence.

\paragraph{Hand pose.}
Hands are represented with MANO~\citep{romero2017mano} and estimated with our in-house annotation pipeline.
The pipeline fits MANO to hand keypoints recovered from the calibrated stereo views, using cross-view reprojection consistency to refine the estimates.
The output is 21 joints per hand in the stereo-left camera frame.
When a hand is outside the view or occluded, no estimate is produced for that frame.

\paragraph{Body motion estimation.}
We estimate the wearer's full-body motion with EgoAllo~\citep{yi2025egoallo}, using the SLAM-derived head trajectory as conditioning and our hand estimates as guidance.
The motion is represented with SMPL-H~\citep{romero2017mano}, with body shape held constant within each recording.
The output comprises 52 body and finger joints at 30\,Hz in the recording's SLAM frame.
The estimated motion is constrained by the observed head and hands, while unobserved body parts, particularly the legs, are inferred using the learned motion prior.

\paragraph{Hierarchical activity annotations.}
We use a large multimodal model to generate hierarchical activity annotations from the recordings.
The annotations describe the recorded activity at three temporal scales.
For the full recording, they summarize the main task and provide an overall activity description.
Within the recording, they identify sub-tasks and actions with associated time intervals.
For each sampled frame, they describe visible objects and their states, the state of each hand, and ongoing hand--object interactions.
All annotations are associated with the video timeline.
The temporally localized actions are distinct from the task-level action categories in Section~\ref{sec:scenes_tasks}.

Registration can then transform the device trajectory and estimated hand and body poses into the scene frame (Section~\ref{sec:spatial_registration}), while the activity annotations remain associated with their corresponding timestamps.

\subsection{\texorpdfstring{Scan--\homie}{Scan--HOMIE} Registration}
\label{sec:spatial_registration}

The scan reconstruction and each \homie recording are expressed in separate coordinate frames.
To place human activity in its spatial context, we register the recording's device trajectory and estimated hand and body motion to the scene frame.
As illustrated in Figure~\ref{fig:viewer}, this brings the egocentric observations, estimated motion, and scene geometry into a shared spatial reference.
Connecting egocentric activity to scene geometry is also central to EPIC Fields, which augments activity videos with reconstructed camera information~\citep{tschernezki2023epicfields}.

We localize stereo-left frames in the scan using SIFT~\citep{lowe2004sift} matching, RANSAC~\citep{fischler1981ransac}, and PnP against scan image keypoints lifted onto the mesh, then fit the recording-to-scene transform by robust Umeyama alignment~\citep{umeyama1991} of the corresponding camera centers.
The transform allows a small scale correction when supported by the trajectory and validation; otherwise, it is rigid.
Held-out frames, IMU gravity consistency, and an ICP~\citep{besl1992icp} check of stereo depth against the mesh are used to assess the alignment and assign PASS, WEAK, or FAIL labels (Appendix~\ref{appendix:registration}).
A single transform cannot correct local SLAM drift within a recording.

\FloatBarrier

%% file: tabs/asset_provenance.tex
\begin{table}[tbp]
\centering
\caption{\textbf{Processed assets and their provenance.}
\emph{Observed} assets are sensor recordings, \emph{estimated} assets are computed from them by geometric or learned methods, and \emph{generated} assets are produced by a language model.
The scene frame is the metric frame of the 3D scan; each \homie recording has its own SLAM world frame until it is registered.}
\label{tab:asset_provenance}
\begingroup
\footnotesize
\renewcommand{\arraystretch}{1.15}
\begin{tabular}{@{}lllll@{}}
\toprule
\textbf{Asset} & \textbf{Device} & \textbf{Frame} & \textbf{Rate} & \textbf{Provenance} \\
\midrule
Fisheye images & 3D scanner & Scene & Per scan & Observed \\
Camera poses, mesh, 3DGS & 3D scanner & Scene & Per scan & Estimated \\
\midrule
Fisheye video ($\times$4), audio, IMU & \homie & Device & 30\,fps / 200\,Hz & Observed \\
Device pose, sparse map & \homie & Recording world & 30\,Hz & Estimated \\
Stereo depth, confidence & \homie & Stereo-left camera & 30\,Hz & Estimated \\
Scene registration $\mathbf{S}$ & Both & Recording $\rightarrow$ scene & Per recording & Estimated \\
Hand pose (MANO) & \homie & Stereo-left camera & 30\,Hz & Estimated \\
Body pose (SMPL-H) & \homie & Recording world & 30\,Hz & Estimated\textsuperscript{$\dagger$} \\
Task, sub-task, and action captions & \homie & --- & Segments & Generated \\
Objects, hand states, narration & \homie & --- & 2\,fps & Generated \\
\bottomrule
\end{tabular}
\endgroup

\vspace{2pt}
{\scriptsize \textsuperscript{$\dagger$}The legs are not observed; the lower-body pose is inferred from head and hand motion.}
\end{table}

%% file: secs/05_dataset_statistics_quality.tex
\section{Dataset Statistics and Quality}
\label{sec:dataset_statistics_quality}


\subsection{Composition and Scale}
\label{sec:composition_scale}

Table~\ref{tab:release_summary} distinguishes the released scan component from the broader collection inventory.
A released scene is one completed scan; duplicate uploads are counted once, while separate scans of the same physical space remain separate scenes.
The processing pipeline supports scene--recording association and geometric registration (Section~\ref{sec:spatial_registration}).
Location and behavior counts describe the broader collection inventory.

\input{tabs/release_summary}

\subsection{Scene QC and Release Decisions}
\label{sec:scene_distribution}

\begin{figure}[!htbp]
  \centering
  \includegraphics[width=\linewidth]{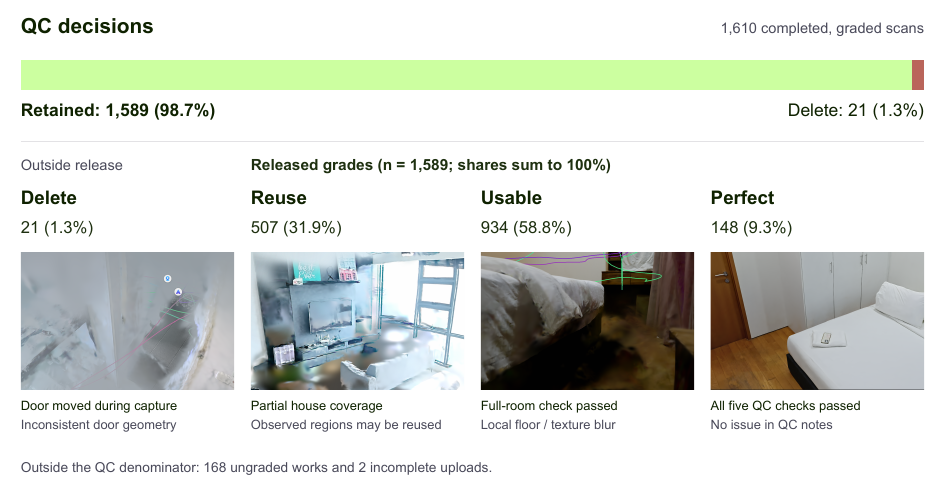}
  \caption{\textbf{QC decisions and released-scene composition.}
  Of \ReleaseDecided{} completed, graded scans, \ReleaseScenes{} are retained (\ReleaseRetainedPercent{}) and \ReleaseDelete{} are Delete (\ReleaseDeletePercent{}).
  Within the retained set, Perfect, Usable, and Reuse account for \ReleasePerfectPercent{}, \ReleaseUsablePercent{}, and \ReleaseReusePercent{}, respectively, summing to 100\%.
  QC-viewer examples and their recorded findings illustrate each grade.}
  \label{fig:scene_qc_flow}
\end{figure}

The upload inventory contains \ReleaseUploadWorks{} works.
Of the \ReleaseDecided{} completed scans with a release grade or Delete decision, \ReleaseScenes{} (\ReleaseRetainedPercent{}) are retained and \ReleaseDelete{} (\ReleaseDeletePercent{}) are Delete (Figure~\ref{fig:scene_qc_flow}).
Another \ReleaseUngraded{} works without an accepted release grade and \ReleaseIncomplete{} incomplete uploads are outside this decision denominator.

Within the \ReleaseScenes{} retained scans, Perfect accounts for \ReleasePerfect{} (\ReleasePerfectPercent{}), Usable for \ReleaseUsable{} (\ReleaseUsablePercent{}), and Reuse for \ReleaseReuse{} (\ReleaseReusePercent{}).
These three grades sum to the released set and are the denominator for release-composition percentages; Delete remains outside the release.
Grades follow the current QC records and release-policy overrides described in Appendix~\ref{appendix:scene_qc}.

The \ReleaseReuse{} Reuse scans include partial coverage and other reconstruction defects.
Reuse denotes potential use of the observed regions, not verified full-room coverage.

The current \ReleaseDelete{} Delete decisions comprise \ReleaseDeleteDoors{} scans with door motion, \ReleaseDeleteFailed{} with failed reconstructions, and \ReleaseDeletePeople{} with moving people or vehicles, according to their linked QC notes.
Release grades also incorporate policy decisions: \ReleaseRuleUsable{} scans lacking an explicit QC result were retained as Usable, and \ReleaseConflicts{} scans with conflicting QC results use the best recorded grade (Appendix~\ref{appendix:scene_qc}).
These assignments are not a new visual reassessment of every reconstruction.

\begin{figure}[!htbp]
  \centering
  \includegraphics[width=\linewidth]{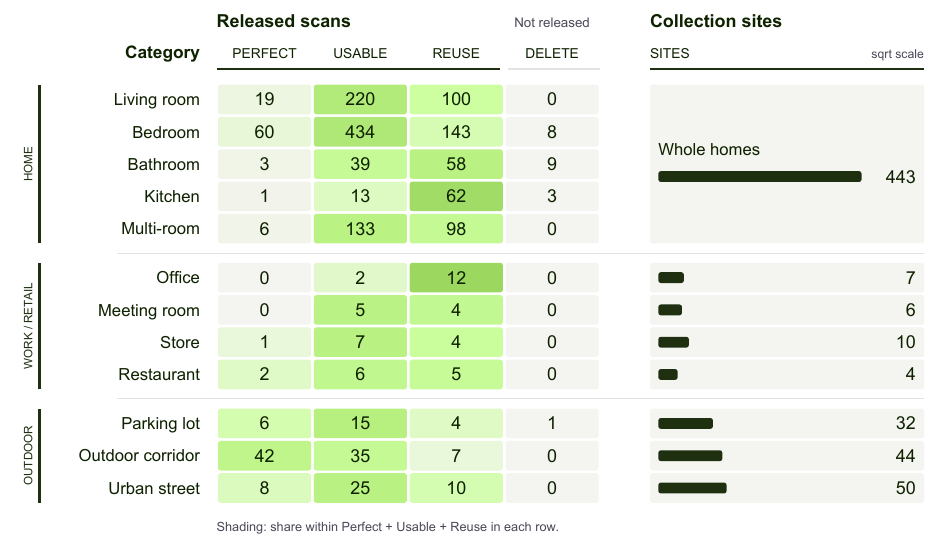}
  \caption{\textbf{Released scene grades, Delete decisions, and collection sites.}
  Left: counts in 12 categories; shading for Perfect, Usable, and Reuse encodes the share within their row sum, excluding Delete.
  Delete counts are shown separately.
  Right: the \CollectionLocations{} sites in the broader collection inventory, on a square-root scale; \CollectionHomes{} homes span all home categories.
  Per-category totals are in Table~\ref{tab:scene_category_statistics}.}
  \label{fig:scene_statistics}
\end{figure}

Figure~\ref{fig:scene_statistics} breaks the counts down by category.
Home interiors account for \ReleaseHomeScenes{} scans (\ReleaseHomePercent{}), and bedrooms and living rooms together for \ReleaseBedroomLivingPercent{} of the release.
Among home categories, kitchens (\ReleaseKitchenReusePercent{}) and bathrooms (\ReleaseBathroomReusePercent{}) have the highest Reuse shares, against \ReleaseReusePercent{} overall.
Among the \ReleaseOutdoorScenes{} outdoor scans, \ReleaseOutdoorGoodPercent{} are Usable or Perfect; outdoor corridors contribute \ReleaseCorridorPerfect{} of the \ReleasePerfect{} Perfect scans.
The independently counted collection inventory contains \CollectionLocations{} locations, including \CollectionHomes{} homes; these site counts are not restricted to the released scans.

\FloatBarrier

%% file: tabs/release_summary.tex
\begin{table}[tbp]
\centering
\caption{\textbf{Released scans and collection inventory.}
The scene count refers to the released scan component.
Locations, tasks, recordings, and duration describe the broader collection inventory.}
\label{tab:release_summary}
\begingroup
\footnotesize
\renewcommand{\arraystretch}{1.15}
\begin{tabularx}{\textwidth}{@{}lXrl@{}}
\toprule
\textbf{Unit} & \textbf{Definition} & \textbf{Count} & \textbf{Breakdown} \\
\midrule
Location & A home or another independent capture site & \CollectionLocations & Figure~\ref{fig:scene_statistics}, right \\
Scene (Level~1) & One released 3D scan and its reconstruction & \ReleaseScenes & Figures~\ref{fig:scene_qc_flow} and~\ref{fig:scene_statistics} \\
Task (Level~2) & A task definition with a goal and instructions & \CollectionTasks & Figure~\ref{fig:task_statistics}a--b \\
Recording & One uploaded \homie recording of one task & \CollectionRecordings & Figure~\ref{fig:task_statistics}c \\
Duration & Total \homie collection time & \CollectionHours\,h & \textemdash \\
\bottomrule
\end{tabularx}
\endgroup
\end{table}

%% file: secs/06_applications.tex
\section{Applications}
\label{sec:applications}

The applications comprise three benchmarks (depth and geometry, 3D object detection, and spatial intelligence) and downstream demonstrations.
All three benchmarks use the same 20 scanned scenes, listed in Appendix~\ref{appendix:benchmark_scenes}.

\subsection{Benchmarking Existing Models}
\label{sec:application_benchmark}

\subsubsection{Depth and Geometry}
\label{sec:application_geometry}
Understanding human activity in a scene requires recovering the surrounding geometry and the observer's motion.
We evaluate how well existing models recover depth, camera trajectories, and scene surfaces from multi-view images, and how their performance changes with the number of input views.

\paragraph{Models.}
The comparison follows SpatialBench~\citep{peng2026spatialbench}, with models grouped into its optimization-based, end-to-end feed-forward, online, chunk-wise, SLAM-based, and test-time-training categories (Table~\ref{tab:application_geometry}).

\paragraph{Metrics.}
All models are evaluated on the same 20 scenes under Sparse (5 frames), Medium (11--100 frames), and Dense (335--500 frames) settings.
Table~\ref{tab:application_geometry}(a) reports depth AbsRel, RMSE, and $\delta_{1.03}$; panel (b) reports pose AUC@$30^\circ$, trajectory ATE, Chamfer distance (CD), and surface F-score at 5\,cm.
The scan reconstructions provide the reference geometry and camera poses.
Depth is median-scaled per scene, so its scores evaluate scale-normalized depth quality.
Frame selection, alignment, and metric definitions are detailed in Appendix~\ref{appendix:application_metrics}.

\paragraph{Results.}
No single model leads across all metrics.
$\pi^3$ achieves the lowest trajectory ATE and the highest surface F-score across all three input settings, while DA3-Giant achieves the highest pose AUC in the Medium and Dense settings.
More input views do not consistently improve performance: CUT3R's pose AUC decreases from 0.4103 to 0.0625 between Sparse and Dense, whereas DA3-Giant's increases from 0.7540 to 0.8074.
Dense inputs also expose memory limitations, with DUSt3R and MASt3R unable to complete evaluation.
These results highlight differences in geometric accuracy and scalability that are not captured by depth scores alone.

\input{tabs/application_geometry}

\subsubsection{3D Object Detection}
\label{sec:application_detection}

\paragraph{Annotations.}
Annotators manually label the 20 evaluation scenes for this benchmark, marking the category and 3D box of each object in the scene frame; these labels are the ground truth.

\paragraph{Models.}
The comparison covers three groups (Table~\ref{tab:application_detection}): end-to-end models that predict boxes from images or point clouds (SpatialLM~\citep{mao2025spatiallm}, LLaVA-3D~\citep{zhu2024llava3d}, SceneScript~\citep{avetisyan2024scenescript}, Map-Det3D~\citep{yang2026mapdet3d}), language fields embedded in a 3DGS (M3~\citep{zou2025m3}, LangSplat~\citep{qin2024langsplat}), and Boxer~\citep{detone2026boxer}, which lifts open-vocabulary 2D detections to 3D boxes.
Each method is queried with the object categories present in the scene.

\paragraph{Metrics.}
AP$_{25}$ and AP$_{50}$ are average precision at 3D IoU thresholds of 0.25 and 0.5; mIoU@0.25 is the mean 3D IoU of predictions matched to a reference box at IoU 0.25.
Query and box conventions are in Appendix~\ref{appendix:application_metrics}.

\paragraph{Results.}
Boxer leads on all three metrics.
The other methods stay below 7 AP$_{25}$, and LLaVA-3D and SceneScript produce no box that reaches IoU 0.25.
No method exceeds 7 AP$_{50}$, so tight 3D boxes in real, cluttered scenes remain beyond current models.

\input{tabs/application_grounding}

\subsubsection{Spatial Intelligence}
\label{sec:application_si}

MMSI-Bench~\citep{yang2026mmsi} provides the reference model roster.
Table~\ref{tab:application_mmsi} groups 500 question--answer pairs from the 20 evaluation scenes into translation, rotation, distance, depth, height, and relative-position questions.
Each group uses answer accuracy (\%), with overall accuracy computed over all questions.

\input{tabs/application_mmsi}

\paragraph{Results.}
Gemini-2.5-Pro reaches the highest overall accuracy (74.2\%); HoliWorld-VL-4B reaches 67.8\%, above every open-weight model (at most 60.4\%) and GPT-5 High (65.8\%).
Every model answers at least 74\% of the height and relative-position questions correctly, whereas none exceeds 34\% on distance or 55\% on translation.

\FloatBarrier
\subsection{Demonstrations}
\label{sec:application_demos}

Generalist robot policies learn visuomotor control from diverse demonstrations, as explored by RT-1 and RT-2~\citep{brohan2022rt1,brohan2023rt2}; Open X-Embodiment provides a cross-robot data collection for such learning~\citep{openx2023embodiment}.
OpenVLA~\citep{kim2024openvla} and the $\pi_0$ and $\pi_{0.5}$ models~\citep{black2024pi0,physicalintelligence2025pi05} further study adapting vision-language priors to robot control and generalization.

The demonstrations use metric, gravity-aligned scene geometry and the pipeline's ability to register \homie recordings to the corresponding scenes.
Real-to-Sim turns a scene into a simulator scene, in which an operator can teleoperate a robot live (virtual teleoperation) and a robot can run new multi-step tasks (long-horizon generation); retargeting drives a robot with recorded human motion.
Each demonstration lists its input, output, and the checks a valid result must pass.

\begin{figure}[!htbp]
\centering
\includegraphics[width=\linewidth]{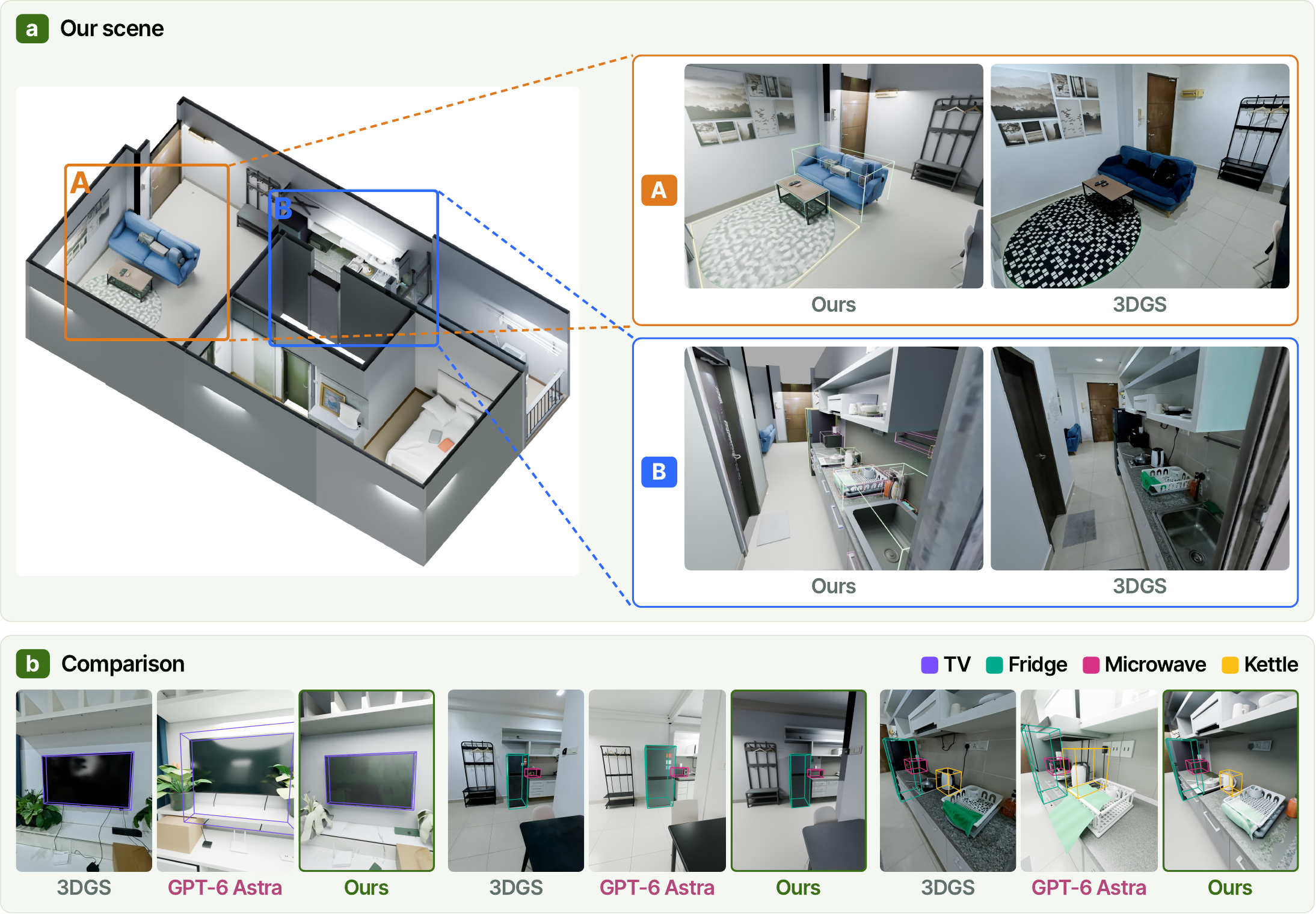}
\caption{\textbf{Real-to-sim conversion of a scanned scene.}
(a) The converted simulator scene, with zoom-ins A and B rendered from the same camera in our scene, with the 3D box of each object, and in the scan's 3DGS.
(b) Three further views rendered from the 3DGS, from a scene built by GPT-6 Astra~\citep{openai2026gpt6astra} from images without 3DGS input, and from our scene, with the 3D boxes of four matched objects.}
\label{fig:application_real2sim}
\end{figure}

\subsubsection{Real-to-Sim}
\label{sec:application_real2sim}
Neural scene representations such as NeRF~\citep{mildenhall2020nerf} support novel-view rendering, while Habitat~2.0 and BEHAVIOR-1K provide environments for physics-based robot interaction~\citep{szot2021habitat2,li2024behavior1k}.
RialTo explores constructing digital twins from real-world captures to improve robot policies~\citep{torne2024rialto}.

\emph{Input:} a captured scene with metric geometry and appearance.
\emph{Output:} a simulator scene and an agent or camera rollout.
\emph{Checks:} scale, gravity alignment, collision geometry, and visual correspondence.
Figure~\ref{fig:application_real2sim}(a) shows a converted scene of 217 separate objects.
Because the scene is built in the scan's frame, any camera can render both the simulator scene and the 3DGS, so the two can be compared view by view without a separate registration step; the zoom-ins A and B match in layout, object extent, and appearance.
Panel (b) repeats the comparison for a scene built from the captured images alone.
Without metric geometry, that scene changes object colours, sizes, and positions, and the boxes of the four matched objects drift from those in the 3DGS.
The boxes measure consistency with the 3DGS rather than accuracy against an independent ground truth.

\begin{figure}[!htbp]
\centering
\includegraphics[width=\linewidth]{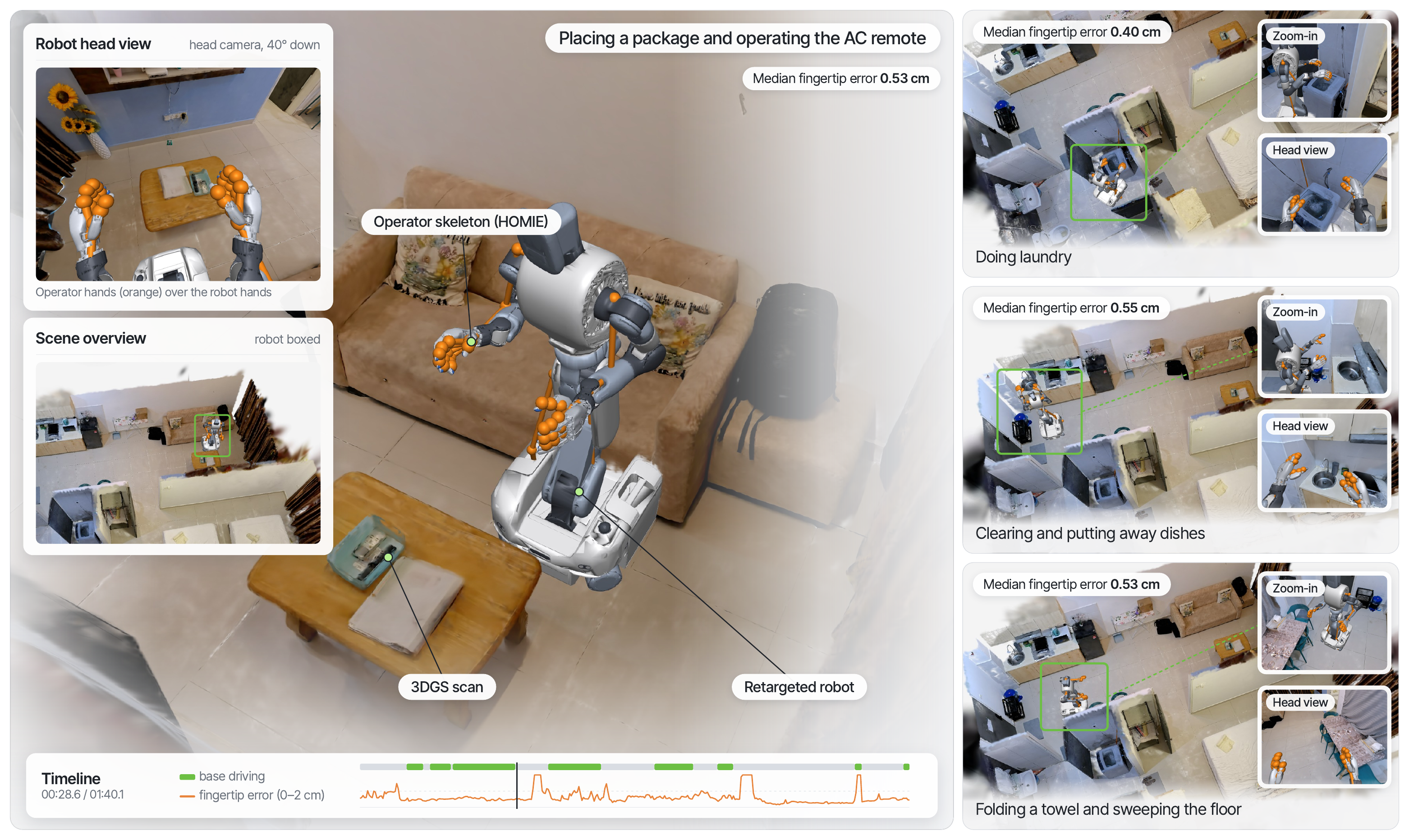}
\caption{\textbf{Human-robot retargeting in a 3D scan.}
\emph{Left:} one frame of a \homie recording retargeted to a wheeled humanoid, with the robot head view, a scene overview, and a timeline of base driving and per-frame fingertip error (clipped at 2\,cm).
\emph{Right:} three further recordings in the same scene, each showing the scene overview with the robot boxed, a third-person zoom-in, and the robot head view.
Tags give the median fingertip error over each recording.
The orange skeleton is the operator; the grey body is the robot. Head views are pitched 40$^\circ$ below the head direction.}
\label{fig:application_retargeting}
\end{figure}

\subsubsection{Human-Robot Retargeting}
\label{sec:application_retargeting}
\emph{Input:} a \homie recording with body and hand motion registered to its scanned scene.
\emph{Output:} base, whole-body, and dexterous-hand joint trajectories of a wheeled humanoid, replayed in the scene's Gaussian reconstruction.
\emph{Checks:} joint velocity and torque within 90\% of their limits, ZMP inside the support polygon, no scene or self-collision, and head and fingertip tracking error.
Recordings are slowed down 2$\times$ before retargeting.
Figure~\ref{fig:application_retargeting} shows four full recordings in one scene, 10.3\,min of motion in total.
The robot drives 96\,m with its base, moving in 35--41\% of the frames, while its hands track the operator's fingertips with median errors of 0.40--0.55\,cm and 95th-percentile errors below 2\,cm.
Every frame passes every check: peak joint speed and torque reach 90\% and 89\% of their limits, and the ZMP stays at least 0.46\,cm inside the support polygon.

\begin{figure}[!htbp]
\centering
\includegraphics[width=\linewidth]{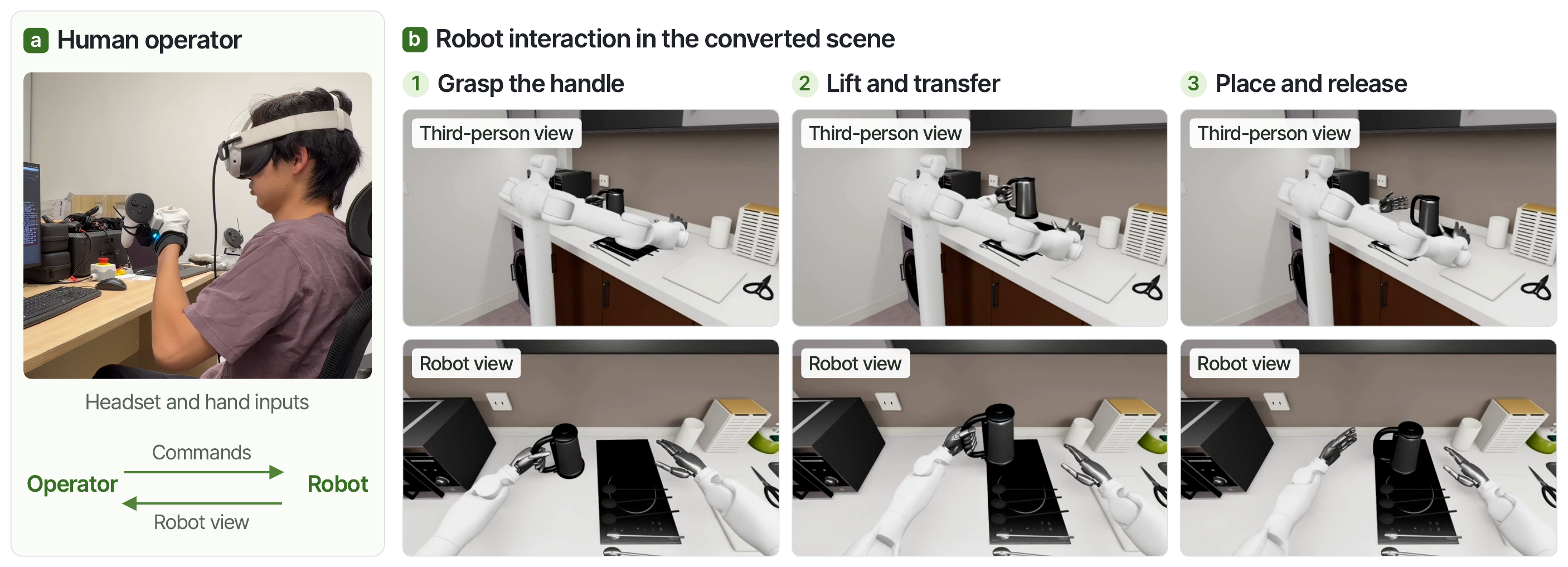}
\caption{\textbf{Virtual teleoperation in a converted scene.}
(a) The operator setup, with a head-mounted display and hand input devices; arrows indicate operator commands and robot-view feedback.
(b) Three stages of a kettle-transfer interaction, each shown from an external camera (top) and the robot camera (bottom): grasping the handle, lifting and transferring the kettle, and placing it on the cooktop and releasing it.
The operator photograph illustrates the setup; the paired simulation views show the interaction stages.}
\label{fig:application_teleop}
\end{figure}

\subsubsection{Virtual Teleoperation}
\label{sec:application_teleop}
\emph{Input:} the simulator scene from Real-to-Sim, built from the scan's mesh and 3DGS, and an operator teleoperating live.
\emph{Output:} a simulated robot under live teleoperation in that scene, and its interaction rollouts.
\emph{Checks:} end-effector tracking of the operator's commands, scene collisions, and joint limits.
Figure~\ref{fig:application_teleop} shows an operator wearing a head-mounted display and hand input devices controlling a simulated humanoid in a converted kitchen scene.
The robot grasps a kettle by its handle, lifts it from the counter, moves it over the cooktop, and places it down before releasing the handle.
The robot-camera and external-camera views show the same interaction from complementary viewpoints.
The resulting rollout contains both robot motion and a change in the object's state, demonstrating that the scanned environment can support the collection of new interactive robot demonstrations.
The scene can be restored in simulation for another collection, without manually resetting a physical setup.

\begin{figure}[!htbp]
\centering
\includegraphics[width=\linewidth]{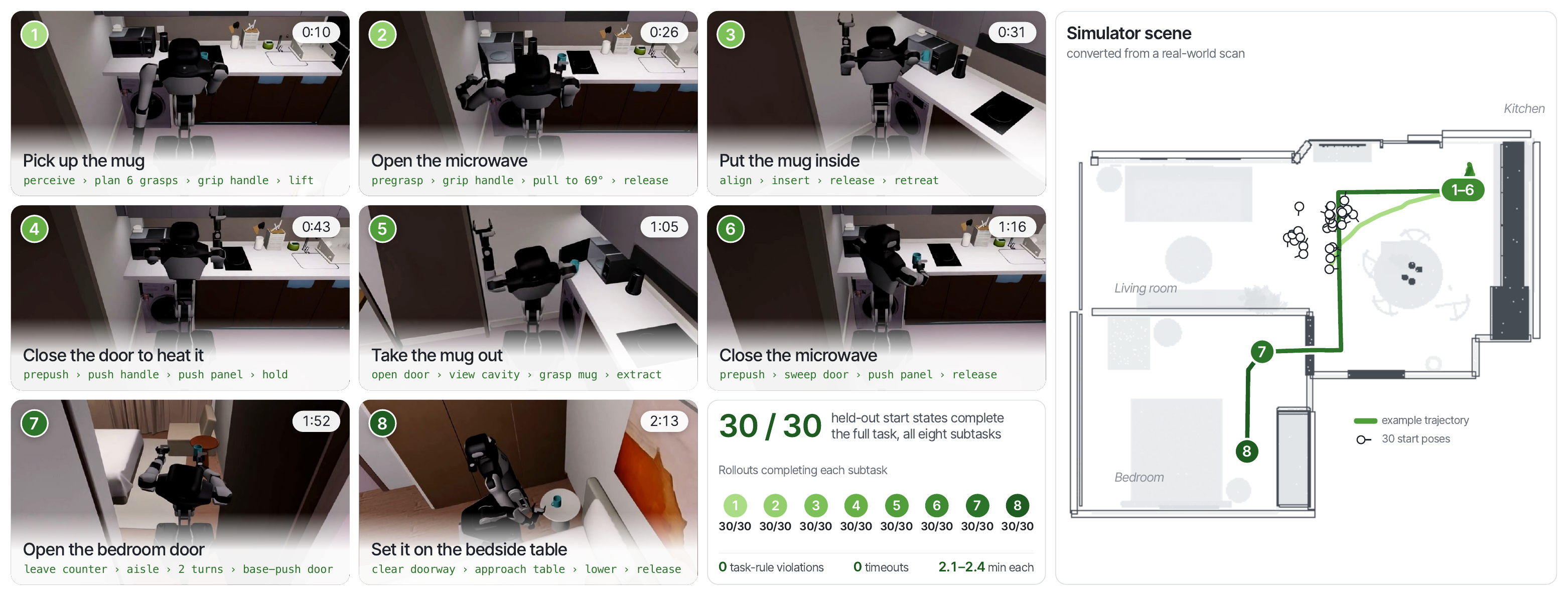}
\caption{\textbf{Long-horizon trajectory generation.} Rollouts of one eight-subtask task in a converted home environment.
\emph{Left:} an example rollout (rendered with a chase camera), with a frame as each subtask completes, the time since the start, and the steps that subtask's policy program logged; the last cell gives the results of the 30 held-out rollouts per subtask and for the full task.
\emph{Right:} the scene's top-down plan with the example base path coloured by subtask, the places where its subtasks end, and the start poses of the 30 held-out rollouts.}
\label{fig:application_long_horizon}
\end{figure}

\subsubsection{Long-Horizon Trajectory Generation}
\label{sec:application_long_horizon}
SayCan and PaLM-E explore language-conditioned task planning~\citep{ahn2022saycan,driess2023palme}, while Code as Policies generates robot policy programs from language~\citep{liang2023codeaspolicies}.
MimicGen studies expanding demonstration datasets by adapting human demonstrations to new contexts~\citep{mandlekar2023mimicgen}.

\emph{Input:} an object-level simulator scene from Real-to-Sim, a robot model, and a task given as a sequence of subtasks.
\emph{Output:} robot trajectories that chain navigation and manipulation through every subtask, from varied initial robot and object poses.
\emph{Checks:} per-subtask and full-task success from the simulator's object states, task rules on the objects (nothing tipped or dropped, appliances and doors left as each subtask requires), trajectory duration, and generation yield and time per valid trajectory.
We generate trajectories for one eight-subtask task in a converted home environment: a wheeled dual-arm robot picks up a mug from the kitchen counter, opens the microwave, puts the mug inside, closes the door to heat it, takes it out, closes the microwave again, pushes the closed bedroom door open, and sets the mug on the bedside table (Figure~\ref{fig:application_long_horizon}).
Each subtask is a policy program that acts through a whole-body controller and observes objects only through perception, not through the simulator state; the eight run in one continuous rollout without resets.
From 30 held-out start states, each with its own robot and mug pose, all 30 rollouts complete all eight subtasks without breaking a task rule.
A trajectory lasts 3,869--4,255 control steps at 30\,Hz (2.2\,min on average) and takes 10--16\,min of wall-clock time on one simulation worker, so further trajectories cost only compute.

\FloatBarrier

%% file: tabs/application_geometry.tex
\begin{table}[!tbp]
\centering
\caption{\textbf{Depth and geometry across input densities.}
Means over the same 20 scenes for Sparse (5 frames), Medium (11--100 frames), and Dense (335--500 frames) inputs, with models grouped by the SpatialBench categories~\citep{peng2026spatialbench}.
Bold marks the best value per metric within each density, including ties; OOM marks out-of-memory runs.
Frame selection, symbols, units, and alignment are described in Appendix~\ref{appendix:application_metrics}.}
\label{tab:application_geometry}
\begingroup
\footnotesize
\setlength{\tabcolsep}{0pt}
\renewcommand{\arraystretch}{1.12}

\makebox[\textwidth][l]{\textbf{(a) Depth}}\par\vspace{3pt}
\begin{tabular}{@{}p{0.20\textwidth}@{}*{9}{>{\centering\arraybackslash}p{\dimexpr0.80\textwidth/9\relax}@{}}}
\toprule
\rowcolor{ropedialightgreen}
\cellcolor{white} & \multicolumn{3}{c}{\textbf{Sparse} (5 frames)} & \multicolumn{3}{c}{\textbf{Medium} (11--100 frames)} & \multicolumn{3}{c}{\textbf{Dense} (335--500 frames)} \\
\cmidrule(lr){2-4}\cmidrule(lr){5-7}\cmidrule(lr){8-10}
Model & AbsRel$\downarrow$ & RMSE$\downarrow$ & $\delta_{1.03}\uparrow$ & AbsRel$\downarrow$ & RMSE$\downarrow$ & $\delta_{1.03}\uparrow$ & AbsRel$\downarrow$ & RMSE$\downarrow$ & $\delta_{1.03}\uparrow$ \\
\midrule
\multicolumn{10}{@{}l}{\emph{Optimization-based}} \\
DUSt3R~\citep{wang2023dust3r} & 0.1485 & 0.6618 & 0.3128 & 0.1758 & 0.6478 & 0.1970 & $\mathrm{OOM}$ & $\mathrm{OOM}$ & $\mathrm{OOM}$ \\
MASt3R$^{\dagger}$~\citep{leroy2024mast3r} & 0.1174 & 0.5946 & 0.2983 & 0.1737 & 0.6264 & 0.2406 & $\mathrm{OOM}$ & $\mathrm{OOM}$ & $\mathrm{OOM}$ \\
\addlinespace[2pt]
\multicolumn{10}{@{}l}{\emph{End-to-end feed-forward}} \\
VGGT~\citep{wang2025vggt} & 0.0833 & 0.4554 & 0.3136 & 0.1043 & \textbf{0.4425} & 0.2435 & 0.1071 & 0.2718 & 0.1978 \\
$\pi^3$~\citep{wang2025pi3} & \textbf{0.0602} & 0.4707 & \textbf{0.5459} & \textbf{0.0727} & 0.5220 & 0.4461 & 0.0666 & \textbf{0.2238} & 0.3855 \\
MAPAnything~\citep{keetha2025mapanything} & 0.0996 & 0.7320 & 0.2829 & 0.1229 & 0.6616 & 0.2107 & 0.1106 & 0.3291 & 0.1861 \\
DA3-Giant~\citep{lin2025depthanything3} & 0.0621 & \textbf{0.4443} & 0.5098 & 0.0854 & 0.5007 & \textbf{0.4723} & \textbf{0.0659} & 0.2475 & \textbf{0.4409} \\
\addlinespace[2pt]
\multicolumn{10}{@{}l}{\emph{Online}} \\
CUT3R~\citep{wang2025cut3r} & 0.1153 & 0.5621 & 0.2024 & 0.1579 & 0.5428 & 0.1505 & 0.1896 & 0.4225 & 0.1083 \\
StreamVGGT~\citep{zhuo2025streamvggt} & 0.1589 & 0.6821 & 0.1945 & 0.1852 & 0.6035 & 0.1252 & 0.1781 & 0.4320 & 0.1144 \\
\addlinespace[2pt]
\multicolumn{10}{@{}l}{\emph{Chunk-wise}} \\
VGGT-Long$^{\ddagger}$~\citep{deng2025vggtlong} & 0.0833 & 0.4554 & 0.3136 & 0.1079 & 0.4479 & 0.2287 & 0.2439 & 0.4917 & 0.1148 \\
\addlinespace[2pt]
\multicolumn{10}{@{}l}{\emph{SLAM-based}} \\
MASt3R-SLAM~\citep{murai2024mast3rslam} & 0.1279 & 0.6028 & 0.2443 & 0.2248 & 0.6767 & 0.0940 & $\mathrm{OOM}^{*}$ & $\mathrm{OOM}^{*}$ & $\mathrm{OOM}^{*}$ \\
\addlinespace[2pt]
\multicolumn{10}{@{}l}{\emph{Test-time training}} \\
TTT3R~\citep{chen2025ttt3r} & 0.1156 & 0.5725 & 0.2202 & 0.1550 & 0.5508 & 0.1471 & 0.1735 & 0.4031 & 0.1250 \\
\bottomrule
\end{tabular}

\vspace{9pt}
\makebox[\textwidth][l]{\textbf{(b) Pose, trajectory, and surface reconstruction}}\par\vspace{3pt}
\begin{tabular}{@{}p{0.20\textwidth}@{}*{12}{>{\centering\arraybackslash}p{\dimexpr0.80\textwidth/12\relax}@{}}}
\toprule
\rowcolor{ropedialightgreen}
\cellcolor{white} & \multicolumn{4}{c}{\textbf{Sparse} (5 frames)} & \multicolumn{4}{c}{\textbf{Medium} (11--100 frames)} & \multicolumn{4}{c}{\textbf{Dense} (335--500 frames)} \\
\cmidrule(lr){2-5}\cmidrule(lr){6-9}\cmidrule(lr){10-13}
Model & AUC$\uparrow$ & ATE$\downarrow$ & CD$\downarrow$ & F$_{5\mathrm{cm}}\uparrow$ & AUC$\uparrow$ & ATE$\downarrow$ & CD$\downarrow$ & F$_{5\mathrm{cm}}\uparrow$ & AUC$\uparrow$ & ATE$\downarrow$ & CD$\downarrow$ & F$_{5\mathrm{cm}}\uparrow$ \\
\midrule
\multicolumn{13}{@{}l}{\emph{Optimization-based}} \\
DUSt3R~\citep{wang2023dust3r} & 0.5963 & 1.4889 & 1.3966 & 0.1160 & 0.4097 & 1.9586 & 2.2121 & 0.1414 & $\mathrm{OOM}$ & $\mathrm{OOM}$ & $\mathrm{OOM}$ & $\mathrm{OOM}$ \\
MASt3R$^{\dagger}$~\citep{leroy2024mast3r} & 0.6545 & 1.5628 & 2.5276 & 0.1285 & 0.4966 & 2.0883 & 2.7135 & 0.1821 & $\mathrm{OOM}$ & $\mathrm{OOM}$ & $\mathrm{OOM}$ & $\mathrm{OOM}$ \\
\addlinespace[2pt]
\multicolumn{13}{@{}l}{\emph{End-to-end feed-forward}} \\
VGGT~\citep{wang2025vggt} & 0.5883 & 0.8538 & \textbf{1.2533} & 0.1031 & 0.6638 & 0.9571 & 1.6356 & 0.1772 & 0.5514 & 1.4556 & 1.9293 & 0.1204 \\
$\pi^3$~\citep{wang2025pi3} & \textbf{0.7838} & \textbf{0.4326} & 2.1282 & \textbf{0.2517} & 0.7727 & \textbf{0.3916} & 2.9932 & \textbf{0.3901} & 0.7793 & \textbf{0.4569} & 3.2861 & \textbf{0.3678} \\
MAPAnything~\citep{keetha2025mapanything} & 0.7145 & 0.6087 & 2.5945 & 0.0978 & 0.6593 & 0.6965 & 2.7199 & 0.2082 & 0.6845 & 0.6674 & 2.5324 & 0.2136 \\
DA3-Giant~\citep{lin2025depthanything3} & 0.7540 & 0.5495 & 1.2892 & 0.1720 & \textbf{0.8486} & 0.3975 & \textbf{1.3618} & 0.3291 & \textbf{0.8074} & 0.4683 & \textbf{1.2913} & 0.3005 \\
\addlinespace[2pt]
\multicolumn{13}{@{}l}{\emph{Online}} \\
CUT3R~\citep{wang2025cut3r} & 0.4103 & 1.7164 & 1.8990 & 0.0431 & 0.2772 & 2.4323 & 2.0217 & 0.0825 & 0.0625 & 2.8776 & 1.5900 & 0.0589 \\
StreamVGGT~\citep{zhuo2025streamvggt} & 0.4873 & 1.7103 & 2.1019 & 0.0629 & 0.4352 & 2.0942 & 1.4756 & 0.0796 & 0.4182 & 2.2610 & 1.8524 & 0.0500 \\
\addlinespace[2pt]
\multicolumn{13}{@{}l}{\emph{Chunk-wise}} \\
VGGT-Long$^{\ddagger}$~\citep{deng2025vggtlong} & 0.5883 & 0.8538 & \textbf{1.2533} & 0.1031 & 0.6142 & 1.1348 & 1.6306 & 0.1634 & 0.2157 & 2.4393 & 1.6160 & 0.0767 \\
\addlinespace[2pt]
\multicolumn{13}{@{}l}{\emph{SLAM-based}} \\
MASt3R-SLAM~\citep{murai2024mast3rslam} & 0.0048 & 3.3117 & 3.2335 & 0.0249 & 0.0049 & 3.3989 & 45.7247 & 0.0100 & $\mathrm{OOM}^{*}$ & $\mathrm{OOM}^{*}$ & $\mathrm{OOM}^{*}$ & $\mathrm{OOM}^{*}$ \\
\addlinespace[2pt]
\multicolumn{13}{@{}l}{\emph{Test-time training}} \\
TTT3R~\citep{chen2025ttt3r} & 0.3712 & 2.3590 & 2.3554 & 0.0425 & 0.2737 & 2.7494 & 2.2366 & 0.0663 & 0.1860 & 2.7832 & 2.1063 & 0.0430 \\
\bottomrule
\end{tabular}
\endgroup
\end{table}

%% file: tabs/application_grounding.tex
\begin{table}[t]
\centering
\caption{\textbf{3D object detection.} Results (\%) on the 20 evaluation scenes; each method is queried with the object categories present in the scene.
Bold marks the best value per metric. $^*$Public checkpoint; boxes are read from its text output.
Metric definitions are in Appendix~\ref{appendix:application_metrics}.}
\label{tab:application_detection}
\begingroup
\small
\setlength{\tabcolsep}{3pt}
\renewcommand{\arraystretch}{1.12}
\begin{tabular}{@{}p{0.16\textwidth}@{}*{7}{>{\centering\arraybackslash}p{\dimexpr0.84\textwidth/7\relax}@{}}}
\toprule
\rowcolor{ropedialightgreen}
\cellcolor{white} & \multicolumn{4}{c}{\textbf{End-to-end models}} & \multicolumn{2}{c}{\textbf{3DGS fields}} & \multicolumn{1}{c}{\textbf{Lifting}} \\
\cmidrule(lr){2-5}\cmidrule(lr){6-7}\cmidrule(l){8-8}
Metric & \makecell[c]{SpatialLM\\\citep{mao2025spatiallm}} & \makecell[c]{LLaVA-3D$^{*}$\\\citep{zhu2024llava3d}} & \makecell[c]{SceneScript\\\citep{avetisyan2024scenescript}} & \makecell[c]{Map-Det3D\\\citep{yang2026mapdet3d}} & \makecell[c]{M3\\\citep{zou2025m3}} & \makecell[c]{LangSplat\\\citep{qin2024langsplat}} & \makecell[c]{Boxer\\\citep{detone2026boxer}} \\
\midrule
AP$_{25}\uparrow$ & 4.23 & 0.00 & 0.00 & 6.45 & 5.72 & 3.43 & \textbf{24.69} \\
AP$_{50}\uparrow$ & 1.07 & 0.00 & 0.00 & 0.13 & 0.64 & 0.34 & \textbf{6.09} \\
mIoU@0.25$\uparrow$ & 43.66 & 0.00 & 0.00 & 36.16 & 37.01 & 37.96 & \textbf{49.48} \\
\bottomrule
\end{tabular}
\endgroup
\end{table}

%% file: tabs/application_mmsi.tex
\begin{table}[!htbp]
\centering
\caption{\textbf{Spatial intelligence.} Answer accuracy (\%) by question type on 500 question--answer pairs from the 20 evaluation scenes; the reference model roster follows MMSI-Bench~\citep{yang2026mmsi}. Bold marks the best value per column.}
\label{tab:application_mmsi}
\begingroup\small
\setlength{\tabcolsep}{4pt}
\renewcommand{\arraystretch}{1.12}
\begin{tabularx}{\textwidth}{@{}l*{7}{>{\raggedleft\arraybackslash}X}@{}}
\toprule
Model & Trans.$\uparrow$ & Rot.$\uparrow$ & Dist.$\uparrow$ & Depth$\uparrow$ & Height$\uparrow$ & Rel.Pos.$\uparrow$ & Overall$\uparrow$ \\
\midrule
\multicolumn{8}{l}{\emph{Open-weight models}} \\
Qwen2.5-VL-7B~\citep{bai2025qwen25vl} & 20.24 & 36.14 & 17.86 & 65.06 & 79.52 & 78.31 & 49.40 \\
Qwen2.5-VL-72B~\citep{bai2025qwen25vl} & 21.43 & 48.19 & 28.57 & 81.93 & 92.77 & 90.36 & 60.40 \\
InternVL3-8B~\citep{zhu2025internvl3} & 25.00 & 30.12 & 19.05 & 68.67 & 95.18 & 80.72 & 53.00 \\
InternVL3-78B~\citep{zhu2025internvl3} & 36.90 & 46.99 & 14.29 & 80.72 & 93.98 & 84.34 & 59.40 \\
LLaVA-OneVision-7B~\citep{li2024llavaonevision} & 15.48 & 24.10 & 13.10 & 56.63 & 85.54 & 74.70 & 44.80 \\
LLaVA-OneVision-72B~\citep{li2024llavaonevision} & 28.57 & 38.55 & 20.24 & 67.47 & 89.16 & 80.72 & 54.00 \\
\multicolumn{8}{l}{\emph{API models}} \\
GPT-4o~\citep{openai2024gpt4o} & 41.67 & 27.71 & 17.86 & 69.88 & 87.95 & 77.11 & 53.60 \\
GPT-5 High~\citep{openai2025gpt5} & 42.86 & 56.63 & 17.86 & 92.77 & 95.18 & 90.36 & 65.80 \\
o3 High~\citep{openai2025o3o4mini} & 38.10 & 62.65 & 11.90 & 85.54 & 93.98 & \textbf{93.98} & 64.20 \\
Gemini-2.5-Pro~\citep{comanici2025gemini25} & \textbf{54.76} & 69.88 & \textbf{33.33} & \textbf{97.59} & \textbf{98.80} & 91.57 & \textbf{74.20} \\
GPT-6 Astra High~\citep{openai2026gpt6astra} & 50.00 & \textbf{72.29} & 27.38 & 91.57 & 97.59 & 92.77 & 71.80 \\
\midrule
HoliWorld-VL-4B~\citep{li2026holiworld} & 51.19 & 62.65 & 19.05 & 87.95 & 96.39 & 90.36 & 67.80 \\
\bottomrule
\end{tabularx}
\endgroup
\vspace{3pt}
\begin{minipage}{\textwidth}
\footnotesize
Overall accuracy pools all 500 questions, of which 23 are multiple-choice and 84 require a numeric answer.
Object names and equivalent direction expressions are normalized; numeric distances are correct within $\max(0.1\,\mathrm{m}, 10\%\times\mathrm{GT})$.
\end{minipage}
\end{table}

%% file: secs/07_release_limitations.tex
\section{Limitations}
\label{sec:known_limitations}

The released scan component is dominated by home interiors and includes Reuse scans with partial coverage or other reconstruction defects.
Release grades incorporate policy overrides and resolution of conflicting QC labels; they should not be interpreted as a uniform visual or geometric validation of all retained scans.
Separate scans may depict the same physical space, and the collection-site inventory is not a verified site census of the released subset.

The recording count and duration describe the collection inventory.
Hand and body motion are estimated rather than motion-capture ground truth, unobserved body parts are inferred, and activity captions are model-generated.
A single recording-to-scene transform cannot correct local SLAM drift.

%% file: secs/08_conclusion.tex
\section{Conclusion}
\label{sec:conclusion}

\datasetname extends Xperience-10M's focus on human behavior with metric scans of real environments and a pipeline for placing egocentric recordings and estimated motion in a shared scene frame.
Its two-stage collection protocol captures environments and behaviors without requiring robots on site.
The resulting scene representations support three benchmarks for evaluating existing models, while real-to-sim conversion, human-robot retargeting, virtual teleoperation, and long-horizon trajectory generation demonstrate how captured scenes and recordings can be reused to produce simulation environments and robot trajectories.
This reuse allows a real capture to support both the study of observed human activity and the generation of further robot experience.
We propose joint capture, spatial registration, and reuse of environments and behaviors as a gold standard for physical AI data collection.
These resources provide a data foundation for studying scaling laws that relate physical AI capability to the scale and diversity of environments and behaviors.

%% file: secs/appendix/a_contributors.tex
\section{Contributors}
\label{appendix:contributors}

\begingroup
\setlength{\parindent}{0pt}
\setlength{\parskip}{5pt}
\textbf{Project Leads:} Hao~Li, Zhaoxi~Chen, Fangzhou~Hong.

\textbf{Contributors:} Yalun~Dai, Runmao~Yao, Kairui~Hu, Ziang~Cao, Mingze~Sun, Xiyuan~Zheng, Yuanyuan~Gao, Xinhao~Ji, Hui~En~Pang, Xuanli~Chen, Dingwen~Zhang, Yap~Kim~Hui, Yichen~Zhao.

\textbf{Data Collectors:} Rui~Duan, Yaobo~Liu, Zetong~Wang, Cheng~Huang~Yean, Jackson~Peng.

\textbf{Advisors:} Ziwei~Liu.

{\footnotesize Names are listed in no particular order.\par}
\endgroup

%% file: secs/appendix/b_processing_annotation.tex
\section{Processing and Annotation Details}
\label{appendix:processing_annotation}


\subsection{Synchronization, Calibration, and Coordinate Frames}
\label{appendix:sync_calibration}

The four \homie cameras are hardware-synchronized to within \SI{50}{\micro\second}.
Video, audio, and IMU samples are timestamped in nanoseconds on the device clock, and every frame-indexed output of the pipeline shares one 30\,fps frame index; the IMU (200\,Hz) is linked to this index through keyframe indices.
Each device is calibrated before capture, with a reprojection error below 0.5\,px on the $1920\times1088$ camera images, and every recording stores the calibration of the device that recorded it: the intrinsics of the four fisheye cameras, the transform from the device body (IMU) frame to the first camera, and the chain of transforms between consecutive cameras.
The two forward-facing fisheye cameras are rectified into a pinhole stereo pair of $1920\times1088$ pixels, of which a $1452\times824$ crop without black borders is used; stereo depth is stored at $480\times272$.

Transforms use homogeneous column vectors, and $\mathbf{T}_{AB}$ maps a point from frame $B$ to frame $A$.
The SLAM world frame of a recording is gravity-aligned with $+z$ up, in meters, with its origin at the head position of the first frame; device poses map the body (IMU) frame to this world frame.
Hand joints are expressed in the rectified stereo-left camera frame (OpenCV convention), and body joints in the SLAM world frame.
After registration (Appendix~\ref{appendix:registration}), the scene frame of the 3D scan replaces the SLAM world frame for all three.

\subsection{Pose and Language Annotation}
\label{appendix:pose_language}

\paragraph{Hands.}
For each hand and frame, our in-house annotation pipeline provides 21 3D joints, the wrist pose, the finger-joint rotations, and the MANO shape coefficients, all in the rectified stereo-left camera frame.
Frames in which a hand is not estimated are marked as missing.

\paragraph{Body.}
The body estimate comprises 52 SMPL-H joints covering the body and fingers, the local joint rotations, the root and head poses in the SLAM world frame, shape coefficients that are constant over the recording, and soft contact scores.
Whenever a hand is observed, the wrist and finger joints of the body agree with its MANO estimate.

\paragraph{Language.}
We use a large multimodal model to generate hierarchical language annotations for each recording.
These annotations summarize the overall task, identify temporally localized sub-tasks and actions, and describe objects, hand states, and hand--object interactions.

\subsection{Scene--Activity Registration}
\label{appendix:registration}

\paragraph{Inputs.}
Registration needs the scan images with their COLMAP camera model and the mesh, and from \homie the SLAM trajectory, the rectified stereo-left video with its intrinsics, and the calibration.
Scans exported only as 3DGS or point clouds cannot be registered.
The recording--scan pairs come from the collection metadata (Section~\ref{sec:capture_protocol}), and the initialization below also verifies that both show the same place.

\paragraph{Global initialization.}
We select well-textured stereo-left frames spread over the recording and match each against the scan images with SIFT features, scoring candidate images by the inliers of a fundamental matrix estimated with RANSAC.
Matched scan keypoints are lifted to 3D by intersecting their camera rays with the mesh, and PnP gives the pose of the \homie camera in the scene frame.
Initialization succeeds when several frames yield a consistent transform, and the initial rotation is the rotation average of their PnP poses.

\paragraph{Localization and fitting.}
About 200 query frames, evenly spaced along the recording, are localized by PnP against scan images retrieved near the pose predicted by the current transform, and localization is repeated once with the fitted transform.
$\mathbf{S}$ is fit by robust Umeyama alignment between the SLAM camera centers and the PnP camera centers, preferring frames with many PnP inliers.
The fit uses even frames, and odd frames give the held-out error.
The scale of $\mathbf{S}$ absorbs small discrepancies between the two metric reconstructions.
A similarity transform is chosen only when the trajectory spans at least 2.5\,m and its held-out error is clearly lower than that of a rigid transform; otherwise $\mathbf{S}$ is rigid.
As a drift indicator, the two halves of the recording are also fit separately and compared.

\paragraph{Independent checks and labels.}
We compare the gravity direction from the \homie IMU with the floor normal of the scan mesh, measure the height of the SMPL-H toe joints above the floor, and compare the stereo depth with the scan: its scale relative to the SLAM landmarks, the fraction of depth points that lie on the mesh and their residual, and the displacement of an ICP refinement started from $\mathbf{S}$.
A registration is labeled FAIL if initialization finds no consistent transform, fewer than eight frames are localized, the held-out error has a 90th percentile above 30\,cm, or ICP moves the registration by more than 30\,cm; the last case indicates that the images agree but the geometry does not, as in an identical-looking neighboring unit.
Otherwise it is labeled PASS when the held-out error has a median of at most 5\,cm and a 90th percentile of at most 12\,cm, the gravity deviation is at most $2^\circ$, and ICP moves the registration by at most 10\,cm and $2^\circ$, and WEAK when any of these criteria is missed.
The toe height is reported but does not affect the label.

\paragraph{Limitations.}
Recordings performed in place, with a trajectory span below about 2\,m, do not constrain scale and are registered rigidly.
Localization succeeds less often for frames that look down at the hands or a tabletop, because these views differ strongly from the scanner's viewpoints.
A single transform cannot correct local SLAM drift, which affects long recordings; per-segment correction is left to future work.

\subsection{Scene QC and Release Decisions}
\label{appendix:scene_qc}

The QC checklist has five checks: (1) preparation, whether the scene was prepared as required by the protocol; (2) door state; (3) full room, whether the whole room is covered; (4) gaps and corners; and (5) occluded regions and views.
QC notes identify defects such as moving people, door motion, invalid traversal, double imaging, failed reconstruction, and partial coverage.
Release grades combine the recorded QC result with the release policy below; the checklist is not a claim that every retained scan passes every check.

The approved release retains only completed uploads assigned Perfect, Usable, or Reuse, counting each work once.
Previous Delete decisions in Living room, Multi-room, Office, Meeting room, Store, Restaurant, Outdoor corridor, and Urban street are reassigned to Reuse.
Completed scans without an explicit QC result are retained as Usable only in Living room, Multi-room, Outdoor corridor, Urban street, and Parking lot.
This rule accounts for \ReleaseOverrideUsable{} scans in the QC registry and \ReleaseSupplementUsable{} additional scans documented in the release supplement.
For conflicting recorded grades, the release takes the best in the order Perfect, Usable, Reuse, Delete; \ReleaseConflicts{} retained scans use this rule.
Delete decisions remaining after these rules are outside the release.
The retained-grade percentages use Perfect + Usable + Reuse (\ReleaseScenes{}); the retained-versus-Delete comparison uses that sum plus Delete (\ReleaseDecided{}).
Ungraded and incomplete works are outside both denominators.

%% file: secs/appendix/c_additional_statistics.tex
\section{Additional Statistics}
\label{appendix:additional_statistics}


\subsection{Per-Category Scene Statistics}

Table~\ref{tab:scene_category_statistics} lists the release-grade counts and the separate Delete counts of Figure~\ref{fig:scene_statistics}.
Collection-site counts describe the broader inventory and are not a site census of the released scans.

\input{tabs/appendix/scene_category_statistics}

\FloatBarrier

%% file: tabs/appendix/scene_category_statistics.tex
\begin{table}[htbp]
\centering
\caption{\textbf{Scene categories, release grades, and Delete decisions.}
Released is the sum of Perfect, Usable, and Reuse; Delete is outside the release.
Collection sites describe the full capture inventory, independently of the released scans; the \CollectionHomes{} homes span all five home categories.}
\label{tab:scene_category_statistics}
\begingroup
\footnotesize
\renewcommand{\arraystretch}{1.08}
\begin{tabularx}{\textwidth}{@{}l*{6}{>{\raggedleft\arraybackslash}X}@{}}
\toprule
\textbf{Category} & \textbf{Perfect} & \textbf{Usable} & \textbf{Reuse} & \textbf{Released} & \textbf{Delete} & \makecell[r]{\textbf{Collection}\\\textbf{sites}} \\
\midrule
\multicolumn{7}{@{}l}{\emph{Home}} \\
Living room & 19 & 220 & 100 & 339 & 0 & \multirow{5}{*}{443} \\
Bedroom & 60 & 434 & 143 & 637 & 8 &  \\
Bathroom & 3 & 39 & 58 & 100 & 9 &  \\
Kitchen & 1 & 13 & 62 & 76 & 3 &  \\
Multi-room & 6 & 133 & 98 & 237 & 0 &  \\
\midrule
\multicolumn{7}{@{}l}{\emph{Work / retail}} \\
Office & 0 & 2 & 12 & 14 & 0 & 7 \\
Meeting room & 0 & 5 & 4 & 9 & 0 & 6 \\
Store & 1 & 7 & 4 & 12 & 0 & 10 \\
Restaurant & 2 & 6 & 5 & 13 & 0 & 4 \\
\midrule
\multicolumn{7}{@{}l}{\emph{Outdoor}} \\
Parking lot & 6 & 15 & 4 & 25 & 1 & 32 \\
Outdoor corridor & 42 & 35 & 7 & 84 & 0 & 44 \\
Urban street & 8 & 25 & 10 & 43 & 0 & 50 \\
\midrule
\textbf{All} & 148 & 934 & 507 & 1{,}589 & 21 & 596 \\
\bottomrule
\end{tabularx}
\endgroup
\end{table}

%% file: secs/appendix/d_experimental_details.tex
\section{Experimental Details}
\label{appendix:experimental_details}

\subsection{Evaluation Subset}
\label{appendix:benchmark_scenes}

All three benchmarks use the same fixed subset of 20 released scenes.
Table~\ref{tab:benchmark_scene_inventory} summarizes its environment categories and release grades.

\input{tabs/appendix/benchmark_scene_inventory}

\subsection{Metrics and Reporting}
\label{appendix:application_protocol}
\label{appendix:application_metrics}

\paragraph{Depth and geometry: SpatialBench.}
The depth metrics are AbsRel, SqRel, RMSE, log-RMSE, and $\delta_\tau$ for $\tau\in\{1.03,1.05,1.10,1.25\}$; $\delta_\tau$ is the fraction of valid pixels with $\max(\hat d/d,d/\hat d)<\tau$.
Pose metrics are rotation and translation-direction accuracy at $3^\circ$ and $5^\circ$, and AUC at $3^\circ$, $5^\circ$, $15^\circ$, and $30^\circ$.
Trajectory metrics are ATE and RPE after Sim(3) alignment; surface metrics are bidirectional Chamfer distance and F-score at 0.05\,m~\citep{peng2026spatialbench}.
The three multiview settings apply to the same scene set.
Sparse and Medium select frames with the greedy visible-voxel coverage objective of SpatialBench: Sparse uses 5 frames and 0.10\,m voxels; Medium uses 0.20\,m voxels and a length-adaptive budget of 11--100 frames, and stops once the selection covers 95\% of the voxels visible from all candidate frames, which some scenes do not reach within the budget.
Dense samples the full trajectory at a fixed stride, giving 335--500 frames.
Voxel sizes and frame budgets are our settings.
Table~\ref{tab:application_geometry} groups the models by the SpatialBench categories and uses one row per model in each metric panel, with separate Sparse, Medium, and Dense column groups; scores are not pooled across densities.
Selected frame identifiers and the actual frame count are fixed per scene and setting across models.
The optional Single setting contains one frame and supports depth only.
Depth is evaluated after per-scene median scaling, and table values are unweighted means over the 20 scenes.
Depth and surface references use only valid scan coverage.

In Table~\ref{tab:application_geometry}, pose AUC is reported at $30^\circ$, and RMSE, ATE, and CD are in meters.
Reference depth, COLMAP poses, and a point cloud sampled from the mesh come from the scan reconstruction and serve as pseudo ground truth; point clouds are evaluated with predicted intrinsics and aligned poses.
MASt3R$^\dagger$ uses DUSt3R-style global optimization, VGGT-Long$^\ddagger$ uses 60-frame chunks with 30-frame overlap and no loop closure, and MASt3R-SLAM keeps the last pose on tracking failure.
OOM marks settings that run out of GPU memory on all 20 scenes; OOM$^{*}$ marks MASt3R-SLAM Dense, where 4 of the 20 scenes exceeded host memory during point-cloud fusion, so no 20-scene mean is available.

\paragraph{3D object detection.}
For each scene, the query vocabulary is the set of object categories present in its manual annotations, and every method predicts 3D boxes for these categories in the metric, gravity-aligned scene frame.
AP$_{25}$ and AP$_{50}$ are average precision at 3D IoU thresholds of 0.25 and 0.5.
mIoU@0.25 averages the 3D IoU over predictions matched to a reference box at IoU 0.25; a method with no such match scores 0.
LLaVA-3D is run from its public checkpoint, whose boxes are parsed from the text output.

\Needspace{4\baselineskip}
\paragraph{Spatial intelligence.}
The model roster follows MMSI-Bench~\citep{yang2026mmsi}.
Questions are typed by translation, rotation, distance, depth, height, and relative position; the overall column pools all 500 questions.

\subsection{Configuration Record}

\begin{table}[!htbp]
\centering
\caption{\textbf{Configuration fields accompanying each result block.}}
\label{tab:application_configuration}
\begingroup
\small
\renewcommand{\arraystretch}{1.08}
\begin{tabularx}{\textwidth}{@{}lX@{}}
\toprule
\textbf{Track} & \textbf{Recorded settings} \\
\midrule
All tracks & Model/checkpoint or API version; code revision; resolution; eligible scenes; failures; hardware. \\
Depth and geometry & Sparse/Medium/Dense setting; selected frame IDs, count, and order; supplied priors; depth alignment; valid coverage. \\
3D object detection & Query categories; input images, point cloud, or 3DGS; 2D proposals; box parameterization. \\
Spatial intelligence & Input views; prompt; decoding settings; answer extraction; question counts by type. \\
Virtual teleoperation & Converted kitchen scene; head-mounted display and hand input devices; kettle-transfer interaction; robot and external camera views. \\
\bottomrule
\end{tabularx}
\endgroup
\end{table}

\paragraph{Virtual teleoperation example.}
The demonstration uses one operator and one kettle-transfer interaction in a converted kitchen scene.
The operator photograph in Figure~\ref{fig:application_teleop}(a) illustrates the input setup.
Panel (b) pairs the robot-camera and external-camera views at three corresponding action stages; the operator photograph is a separate setup view.


%% file: tabs/appendix/benchmark_scene_inventory.tex
\begin{table}[htbp]
\centering
\caption{\textbf{Evaluation subset composition.}
Environment categories and quality grades follow the approved release.}
\label{tab:benchmark_scene_inventory}
\begingroup
\small
\renewcommand{\arraystretch}{1.08}
\begin{tabularx}{\textwidth}{@{}Xrrr@{}}
\toprule
\textbf{Environment} & \textbf{Perfect} & \textbf{Usable} & \textbf{Total} \\
\midrule
Indoor & 8 & 7 & 15 \\
Outdoor & 5 & 0 & 5 \\
\midrule
\textbf{All} & \textbf{13} & \textbf{7} & \textbf{20} \\
\bottomrule
\end{tabularx}
\endgroup
\end{table}

%% file: secs/appendix/e_acknowledgments.tex
\FloatBarrier
\section{Acknowledgments}
\label{appendix:acknowledgments}

We acknowledge Ropedia and HESAI for the capture technologies used in this work.
Egocentric recordings were captured with Ropedia's HOMIE Gen2 head-mounted recording system, and scene scans were acquired with HESAI's KOSMO handheld 3D scanner.